\pdfoutput=1

\documentclass[11pt]{article}

\usepackage[]{EMNLP2022}

\usepackage{times}
\usepackage{latexsym}
\usepackage{graphicx}
\usepackage{threeparttable}
\usepackage{multirow}
\usepackage{multicol}
\usepackage{subfigure}
\usepackage[T1]{fontenc}

\usepackage[utf8]{inputenc}

\usepackage{microtype}

\usepackage{inconsolata}

\title{{D$^4$}: a Chinese \textbf{D}ialogue \textbf{D}ataset for \textbf{D}epression-\textbf{D}iagnosis-Oriented Chat}

\author{
  Binwei Yao$^{1,2,3}$, Chao Shi$^{3}$, Likai Zou$^{3}$, Lingfeng Dai$^{1,2,3}$\\ 
  \bf {Mengyue Wu}$^{1,2,3}$$^*$, \bf{Lu Chen}$^{1,2,3}$\thanks{*Mengyue Wu and Lu Chen are the corresponding authors.},  \bf{Zhen Wang}$^{4,5}$ \and \bf{Kai Yu}$^{1,2,3}$
  \\
  $^1$SJTU X-LANCE Lab, Department of Computer Science and Engineering\\
  $^2$MoE Key Lab of Artificial Intelligence, SJTU AI Institute\\
  $^3$Shanghai Jiao Tong University, Shanghai, China\\
  $^4$Shanghai Mental Health Center\\
  $^5$Shanghai Jiao Tong University School of Medicine, Shanghai, China\\
  \texttt{\{yaobinwei, mengyuewu, chenlusz, kai.yu\}@sjtu.edu.cn}\\
  \\
\\
}
\begin{document}
\maketitle
\begin{abstract}

In a depression-diagnosis-directed clinical session, doctors initiate a conversation with ample emotional support that guides the patients to expose their symptoms based on clinical diagnosis criteria. 
Such a dialogue system is distinguished from existing single-purpose human-machine dialog systems, as it combines task-oriented and chit-chats with uniqueness in dialogue topics and procedures.
However, due to the social stigma associated with mental illness, the dialogue data related to depression consultation and diagnosis are rarely disclosed. 
Based on clinical depression diagnostic criteria ICD-11 and DSM-5, we designed a 3-phase procedure to construct D$^4$: a Chinese Dialogue Dataset for Depression-Diagnosis-Oriented Chat\footnote{To get access to D$^4$, please look at \href{https://x-lance.github.io/D4}https://x-lance.github.io/D4.}, which simulates the dialogue between doctors and patients during the diagnosis of depression, including diagnosis results and symptom summary given by professional psychiatrists for each conversation.
Upon the newly-constructed dataset, four tasks mirroring the depression diagnosis process are established: response generation, topic prediction, dialog summary, and severity classification of depressive episode and suicide risk. 
Multi-scale evaluation results demonstrate that a more empathy-driven and diagnostic-accurate consultation dialogue system trained on our dataset can be achieved compared to rule-based bots. 
\end{abstract}

\section{Introduction}
\par Given the increasing worldwide health threat brought by depression, researchers have been exploring effective methods for depression detection and diagnosis. 
Besides automatic depression detection from posts on social media~\citep{orabi2018deep}, speech~\citep{zhang2021depa} and multi-modality~\citep{cummins2013diagnosis}, the dialogue system is considered an effective tool for large-scale depression detection~\citep{pacheco2021smart}. It is believed that conversation agents could reduce the concealment of sensitive information such as suicidal thoughts caused by social expectations~\citep{schuetzler2018influence} and the emotional hindrance due to the pressure of being judged in face-to-face conversation~\citep{hart2017virtual}.
In past research, chatbots initiated for depression diagnosis are generally implemented based on self-rating scales~\citep{jaiswal2019virtual,arrabales2020perla} or diagnostic criteria~\citep{philip2017virtual}. The final diagnosis results are obtained by asking fixed questions on the scale and corresponding the user's answers to each question to the scale options. 
These chatbots present good sensitivity and specificity in diagnosis and are more attractive and acceptable \citep{vaidyam2019chatbots,abd2019overview} than the original self-rating scales.
Nevertheless, the fixed dialogue flow limiting the user's expressions to specific answers can not realize personalized consultation and give emotional support at an appropriate time, for which there still exists a big gap between the conversation experience current depression diagnosis agents provide and the face-to-face interview in the process of clinical diagnosis.
\begin{figure}[tbp]
    \centering
    \includegraphics[width=0.48\textwidth]{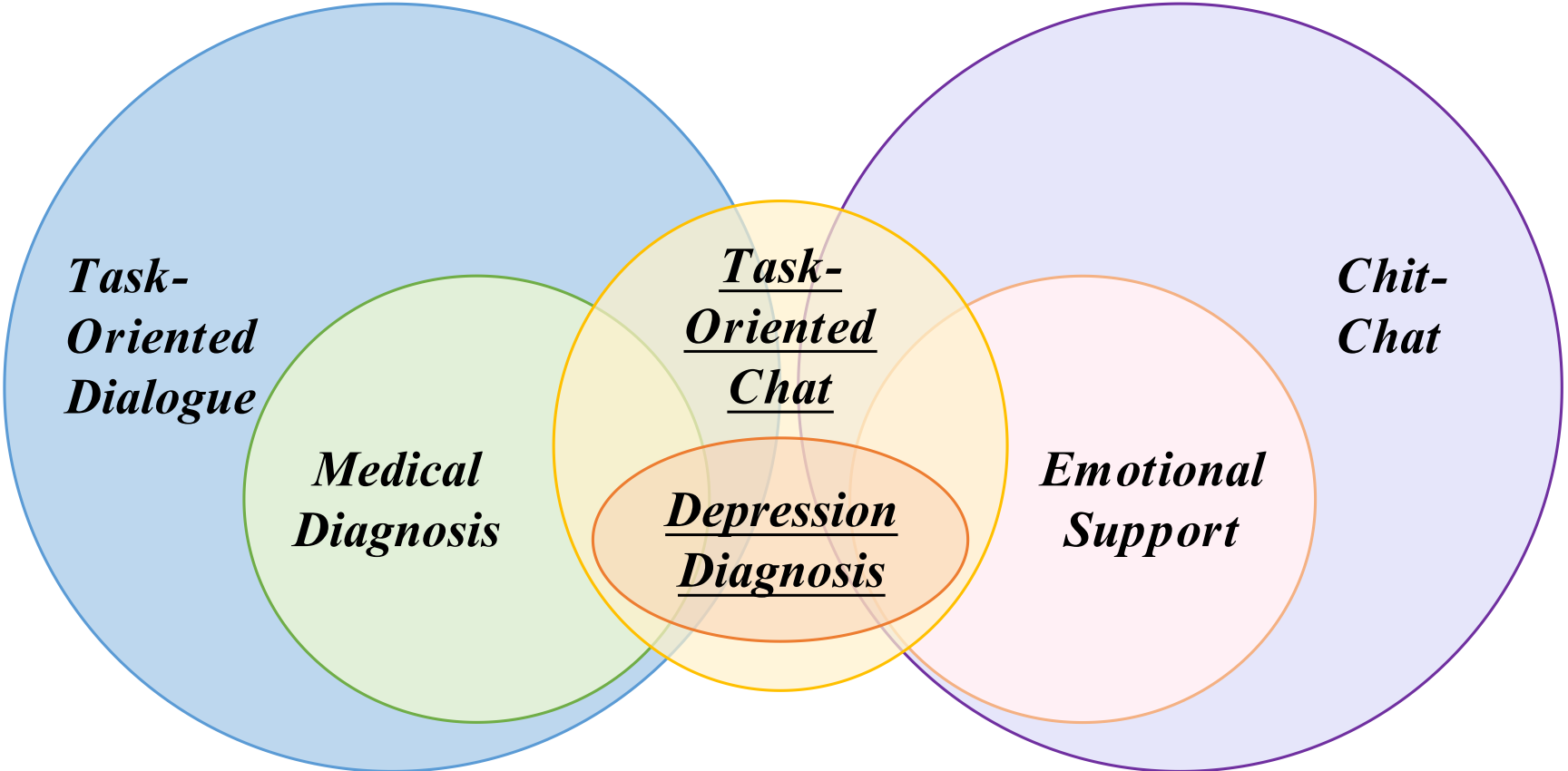}
    \caption{Comparison of Different Dialogue Types}
    \label{fig:dialogue comparison}
\end{figure}
\par Interview-based clinical diagnosis in psychiatry is a complex procedure with the purpose of collecting and summarizing key symptom information about one patient while providing a chat-like conversation experience. In clinical practice, psychiatrists communicate with patients and offer diagnosis results based on practical experience and multiple diagnostic criteria. The most clinically-adopted criteria involve ICD-11~\citep{ICD11:2021}, DSM-5~\citep{APA2013}, etc., which define core symptoms for the depression diagnosis. At the same time, psychiatrists provide emotional support such as empathy and comfort during the consultation to better prompt patients' self-expression. The practice of clinical depression diagnosis displays the possibility of the depression diagnosis dialogue system in further improving the accuracy of diagnosis and user engagement.
\par Accordingly, the depression diagnostic conversation belongs to a distinguished dialogue from previously defined dialogue typologies, which is a combination of task-oriented dialogue and chit-chat. Such a compound dialogue type could be defined as \textit{Task-Oriented Chat} as shown in Figure \ref{fig:dialogue comparison}. This type of dialogue requires multiple assessments regarding task completion and chit-chat experience, which are extremely challenging and still under-investigated. As a specific domain of Task-Oriented Chat, the depression diagnosis dialogue has a clear purpose of the task-oriented dialogue aiming at medical diagnosis: to collect the patient's symptom information and draw a diagnosis conclusion while simultaneously bearing the needs of a chit-chat dialogue with emotional support: to start a user-oriented chat and provide emotional support from time to time.  Currently, no datasets are specified for depression diagnosis, mainly due to the social stigma associated with clinical privacy and the complexity of the diagnosis process. 
\par
\begin{figure*}
    \centering
    \includegraphics[width=\textwidth]{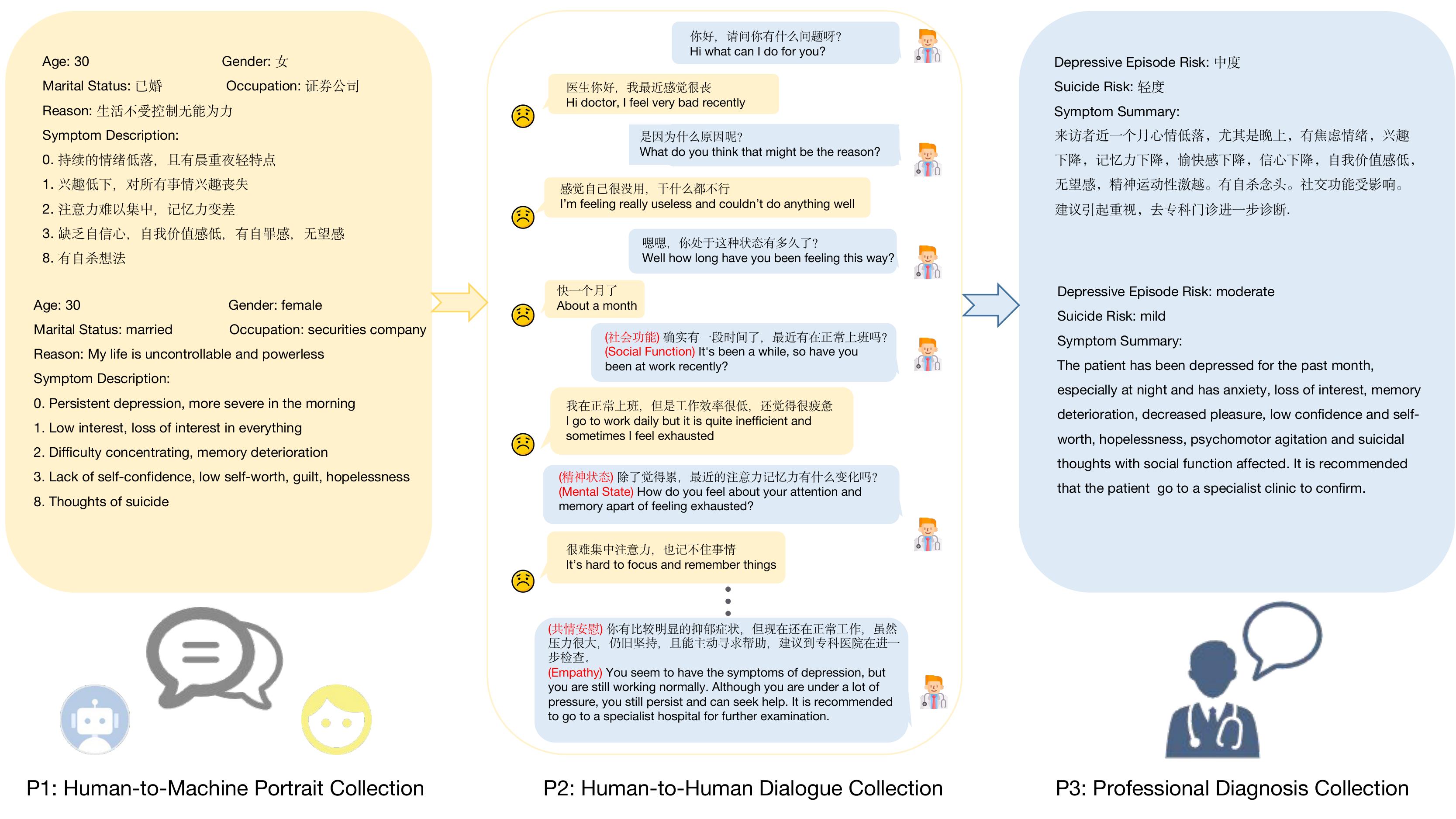}
    \caption{The 3-Phases Data Collection: P1, P2, and P3 denotes the three phases in data collection}
    \label{fig:3 phases}
\end{figure*}

 To construct a clinically sound and empathetic depression-diagnosis-oriented dialogue system close to clinical practice, we conduct dialogue collection through consultation dialogue simulation. We devise a three-phase approach to collect depression diagnostic dialogues (see Figure \ref{fig:3 phases}). \textbf{P1:} To \textit{simulate medical records}, we collect actual patients' portraits with a consultation chatbot web app that asks users fixed questions abstracted from clinical depression diagnosis criteria ICM-11 and DSM-5. \textbf{P2:} To \textit{restore psychiatric consultation conversations}, we employ workers to conduct the consultation dialogue simulation based on the collected portraits. The workers are divided into patients and doctors for separate training by professionals. The doctor actor is required to obtain fixed symptom information involved in the diagnostic criteria in the chat, while the patient actor needs to express according to the symptoms in the portrait. \textbf{P3:} To \textit{reinforce the clinical setting}, professional psychiatrists and psychotherapists supervise the whole process and filter out unqualified dialogues. In addition, they provide diagnosis summaries based on the portrait and dialogue history. We further annotate the conversation procedure with 10 topic tags and the symptom summaries with 13 symptom tags (grouped by core depressive symptoms listed in DSM-5 and ICD-11). In this way, we propose \textbf{D$^4$}: a Chinese \textbf{D}ialogue \textbf{D}ataset for \textbf{D}epression-\textbf{D}iagnosis-Oriented Chat.
The key contribution of this paper is as follows:
\begin{itemize}
    \item A close-to-clinical-practice depression diagnosis dataset with 1,339 conversations generated from actual populations' portraits, accompanied by psychiatrists' diagnosis summaries, under the framework of most applied clinical diagnosis criteria ICD-11 and DSM-5, with multi-dimensional analysis suggesting that our simulated diagnosis data are reliable and up to professional standards.
    \item Experimental validation on four tasks that mirror the real-life diagnosis process: response generation, topic prediction, dialog summary, and severity classification of depression and suicide risk; 
    \item To the best of our knowledge, this is the first diagnosis dialogue dataset for mental health, aiming to advance the realization of an Avante-Garde clinical diagnosis-oriented dialogue system that combines characteristics of task-oriented dialogue and chit-chat.
\end{itemize}

\section{Data Collection}
To maximize doctor and patient authenticity in a diagnosis dialogue, we devise a 3-phase collection paradigm (see Figure\ref{fig:3 phases}) instead of the commonly-adopted vanilla crowdsourcing scheme: \textbf{P1.} We collected natural populations' portraits (in particular actual depressive patients) to form pre-diagnosis records; \textbf{P2.}  Simulated natural diagnostic consultation dialogues based on the portraits; \textbf{P3.} Psychiatrists proofread dialogue history and prescribed professional symptom summaries. 

\subsection{Human-to-Machine Portrait Collection}
To overcome the impracticability in obtaining patients' medical records covered by doctor-patient confidential protocol, we designed a consultation chatbot based on the state machine, which utilizes fixed questions from clinical criteria to document each user's depression symptoms and \textit{demographic information such as age, gender, marital status and occupation}. \textit{Depression symptoms} are prompted accordingly, including \textit{mood, interest, mental status, sleep, appetite, social function, and suicidal tendency}. Users are invited to respond concisely, e.g., yes/no answer and severity estimation. Combined, we obtained a voluntary and legit depression portrait. 
As of the submission of the paper, we have collected a total of 478 patient portraits. We estimate the severity of depressive episodes and suicide risk based on clinical criteria ICD-11 and DSM-5 for each patient portrait . The result is shown in Table \ref{tab: risk estimation of portraits}. Sixty-eight portrait providers reported being diagnosed with depression in an authorized clinic. Among these providers, 53 are currently experiencing a depressive episode.

\subsection{Human-to-Human Dialogue Collection}
To guarantee the quantity, quality, and professionalism of our consultation dialogues, we conducted conversation simulations under the guidance of psychiatrists, 
following portraits collected in Phase 1. 
In particular, we first gathered a small number of dialogues between doctors and patients in real scenarios. Based on the prerequisites mentioned above and clinical depression diagnosis criteria ICD-11 and DSM-5, we released the simulation tasks to crowdsourcing workers. The whole procedure is introduced accordingly: 1) Design and Training: the workers first go through specialized training and are then divided into doctor and patient roles; 2) Annotation: During the conversation, they are required to annotate topic transitions; 3) Peer Assessment: doctor and patient roles rate each other on multiple dimensions after the conversation.
\begin{table}[htbp]
    \centering
    \resizebox{1\linewidth}{!}{
    \begin{tabular}{ccccc}
    \hline
    \textbf{Risk}     &  \textbf{Control} & \textbf{Mild} & \textbf{Moderate} & \textbf{Severe} \\
    \hline
    Depression     & 264&49&95&70\\
    Suicide     & 338&46&75&19\\
    \hline
    \end{tabular}}
    \caption{Risk Estimation of Portraits: "control" represents no risk, "mild", "moderate", and "severe" represent the severity of the risk respectively}
    \label{tab: risk estimation of portraits}
\end{table}
\subsubsection{Design and Training}
\label{ssec:design and training}
\paragraph{Acting Patients} It should be noted that most of our patient actors are not depressive patients. To help them better interpret the symptoms in the patient portraits, we provide detailed explanations, including the severity and duration, and some patients' self-reports to help them understand their inner feelings.
Based on the accurately expressed symptoms, they extend the natural expressions of each aspect following doctors' inquiries in the conversation. 
\paragraph{Acting Doctors} Firstly, we invite licensed psychiatrists and clinical psychotherapists to initiate consultation conversations with actual depressive patients, from which we collect reference conversations. Then based on these essential histories, combined with ICD-11 and DSM-5, we compile 41 symptom items necessary when diagnosing depression and design the questioning logic between questions of symptoms from mild to severe.
The inquiries weren't set as specific expressions for data diversity. Thus, the acting doctors needed to use colloquial rhetoric to ask relevant information involved in these questions and obtain enough information from the patient. 
Meanwhile, to further improve the dialogue experience, we require the acting doctors to conduct a user-oriented dialogue and provide emotional support when necessary.
All acting doctors start the dialogue simulation after completing the training process.

\subsubsection{Topic Annotation}
Considering that the depression diagnostic dialogue has ambiguity between the chat and task-oriented dialogue, it's difficult to define a clear ontology as other task-oriented dialogues\citep{chen2022opal}. To facilitate dialogue generation, we conducted topic annotation on doctors' utterances. According to core symptoms covered in the clinical criteria, we categorized the dialogue topics into \textit{mood, interest, mental status, sleep, appetite, somatic symptoms, social function, suicidal tendency, and screening}. Notably, we included \textit{empathy} as a special topic since it is an essential part of clinical practice. The doctor actors were asked to mark the topics for each utterance during the conversation.
\subsubsection{Peer Assessment}
After the conversation, both sides are required to rate each other in several dimensions for the need for quality control which will be detailed in \ref{ssec: quality control}.
\subsection{Professional Diagnosis Collection}
\label{ssec: Professional Diagnosis Collection}
To ensure the accordance with clinical protocol, we further invite professional psychiatrists and clinical psychotherapists to screen the dialogues that meet the diagnostic standards and provide psychiatric diagnostic results and symptom summaries. At the same time, they score the acting doctors and patients separately with the real-scenario resemblance degree.
\subsection{Quality Control}
\label{ssec: quality control}
Hierarchical screenings are conducted to control the data quality: whether it is up to clinical standard and can satisfy our model training purpose.  
Besides psychiatrists' clinical protocol screening mentioned in part \ref{ssec: Professional Diagnosis Collection}, we adopt a variety of paradigms to conduct quality examinations for better training. We set minimum limits on the length of the dialogue, the average utterance length per dialogue of the doctor, the mutual scores, and the scores given by the psychiatrist shown in the Table \ref{tab:Quality Control Criteria}. The unqualified dialogues are excluded. 

Ultimately, we collected a total of 4,428 conversations and finally retained 1,339 (30\%) after our stringent up-to-clinical-standard quality screenings.
\begin{table}[]
    \centering
    \resizebox{1\linewidth}{!}{
    \begin{tabular}{ccc}
    \hline
    \textbf{Aspects} &  \textbf{Rating Content} & \textbf{Minimum} \\
    \hline
    \multirow{4}{*}{Patient}&expression naturalness & 3(5)\\
    &narrative consistency  &3(5)\\
    &matching extent of symptom  &\multirow{2}{*}{3(5)}\\
    &severity and expression* &\\
    \hline
    \multirow{3}{*}{Doctor}      & degree of similarity to the doctor&3(5)\\
     & degree of similarity to the doctor*&3(5)\\
     & Avg.length of utterances& 8\\
    \hline
    Total & Avg. utterances per dialogue& 30\\
    \hline
    \end{tabular}}
    \caption{Quality Control Criteria: Scores* is given by psychiatrists, the rest are obtained by peer assessment; Numbers in parentheses = the highest score}
    \label{tab:Quality Control Criteria}
\end{table}

\section{Data Characteristics}
\subsection{Statistics}
The overall statistics of the dataset are shown in Table \ref{tab: Statistics of Dataset}. As seen in such a diagnosis scenario, sufficient dialogue turns are required: our diagnosis dialogue exhibit avg. 21.6 turns and avg. 877.6 tokens per dialogue, significantly longer than previous related datasets, suggesting the discrepancies of a diagnosis dialogue task and its distinguished data requirements. 
Meanwhile, our dataset has colloquial and diverse expressions shown by the number of n-grams and avg. 14.4 tokens per utterance.
\begin{table}[]
    \centering
      \centering
    \resizebox{1\linewidth}{!}{
    \begin{tabular}{cccc}
    \hline
    \textbf{Category}     &  \textbf{Total} & \textbf{Patient} & \textbf{Doctor}  \\
    \hline
    Dialogues     & 1339&-&-\\
    Avg. turns     & 21.6&-&-\\
    Workers     & 201 &127&74\\
    Avg. utterances per dialogue & 60.9&30.9&29.9\\
    Avg. tokens per dialogue & 877.6 & 381.8 & 495.8 \\
    Distinct 3-grams of dialogues &245,553 &148,269 &128,203\\
    Distinct 4-grams of dialogues&452,012 &251,121 &224,476\\
    Distinct 5-grams of dialogues&617,233 &324,738 &304,128\\
    Avg. tokens per utterance & 14.4&12.3&16.6\\
    Avg. tokens per symptom summary  & 84.4 & - & - \\
    \hline
    \end{tabular}}
    \caption{D$^4$ Statistics}
    \label{tab: Statistics of Dataset}
\end{table}

\begin{figure*}[hbtp]
    \centering
    \includegraphics[width=\textwidth]{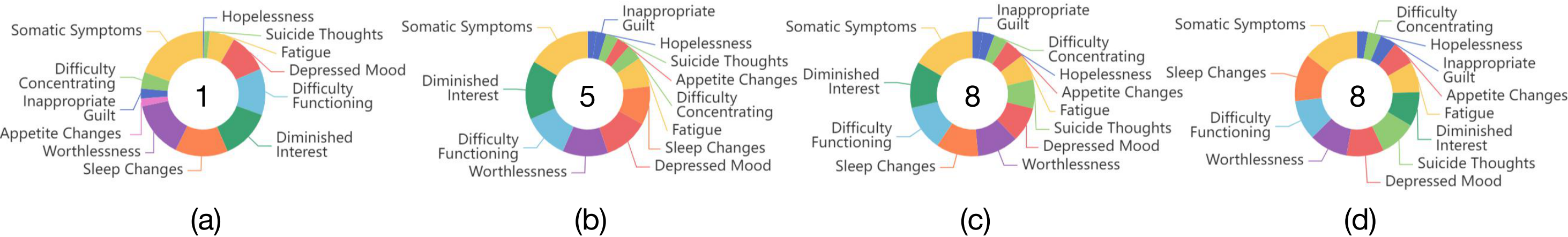}
    \caption{The Symptom Ratio of Summaries: the depressive episode severity increases from (a) to (d) with avg. number of symptoms in the center of each pie chart}
    \label{fig: diagnosis sheets of different depression severity}
\end{figure*}
\subsection{Depression Severity Analysis}
To observe differences in patients with different depression severity, we analyzed conversational and summary symptom statistics by seriousness.
\paragraph{Distribution Feature}
We present statistics on patients' severity of depressive episodes in Table \ref{tab: Statistics of patients of varying severity}. As the degree of depression worsens, the turns and dialog lengths get longer due to doctors' more in-depth questions on specific topics. The diagnostic summaries are also longer to include more symptoms. The most frequent topics are also subject to change with severity: \textit{suicidal tendency} is more likely to be questioned among severer patients.
\begin{table}[htbp]
    \centering
    \setlength\tabcolsep{3pt} 
    \resizebox{1\linewidth}{!}{
    \begin{threeparttable}
    \begin{tabular}{ccccc}
    \hline
    \textbf{Category}     &  \textbf{Control} & \textbf{Mild} & \textbf{Moderate} & \textbf{Severe}  \\
    \hline
    Dialogues     & 430&342&368&199 \\
    Avg. turns     & 17.9&21.3&23.7&26.0 \\
    1st frequent topic &Emp. &Emp.&Emp.&Emp. \\
    2nd frequent topic & MS&MS&MS&Suicide \\
    3rd frequent topic & Sleep &Mood&Suicide&MS \\
    Avg. tokens of symptom summary & 59.8 & 82.0 & 100.5 & 111.9\\
    \hline
    \end{tabular}
     \begin{tablenotes}
        \item \textbf{Emp.}:Empathy \textbf{MS}:Mental Status 
      \end{tablenotes}
  \end{threeparttable}}
    \caption{Depression Severity Statistics in D$^4$}
    \label{tab: Statistics of patients of varying severity}
\end{table}
\paragraph{Analysis of Symptom Summary}
We annotated the 13 core symptoms in the symptom summary according to ICD-11. From Figure \ref{fig: diagnosis sheets of different depression severity}, we observe a difference in the symptom number and ratio from diagnosis summaries of varying severity. As shown in Chart (a), control participants have only a few symptoms, and most are superficial symptoms like sleep changes and worthlessness, commonly in healthy populations. As the condition worsens, the patient has more symptoms, the proportion of each symptom in the summary is gradually averaged, and suicide thoughts become more frequent. The moderate and severe patients share the same average symptom number, indicating that a more fine-grained classification of depression severity requires additional information besides the number of symptoms, such as the duration and severity of each symptom.

\subsection{Topic Analysis}
\begin{figure}[htbp]
    \centering
    \includegraphics[width=0.3\textwidth]{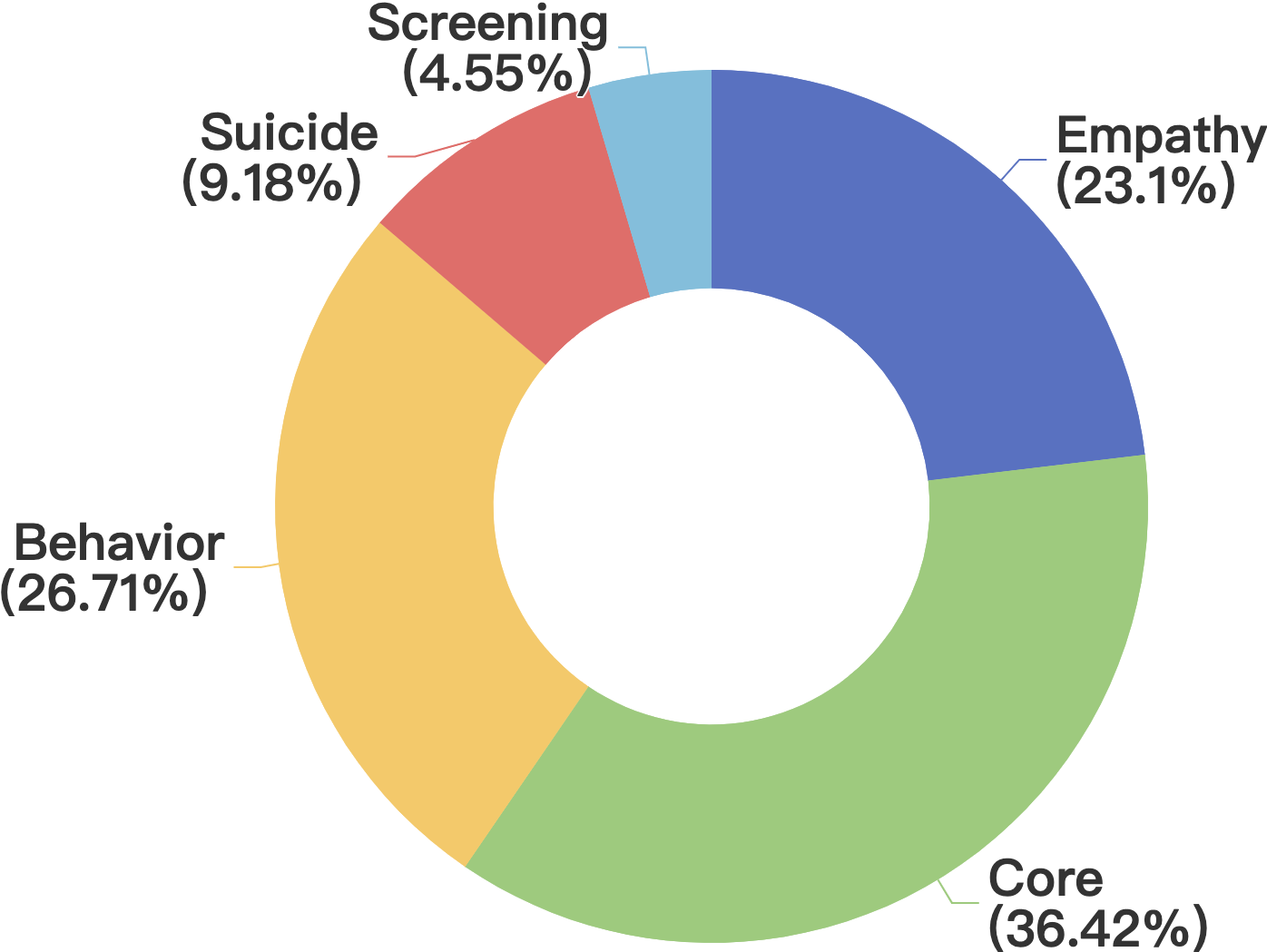}
    \caption{Topic Proportion}
    \label{fig:topic proportion}
\end{figure}

To analyze the characteristics of the doctor's consultation method, we provide perspectives on topic distribution, transition, and lexical features of empathy.
\paragraph{Topic Distribution}
To better analyze the proportion of different symptoms, we regrouped the 10 topics annotated by acting doctors. \textit{mood, interest, mental status, social function} are grouped into \textit{core} and \textit{sleep, appetite, somatic symptom} are grouped into \textit{behavior}. Figure \ref{fig:topic proportion} shows the propotion of regrouped topics. \textit{Core} and \textit{behavior} occupy 63.17\% of the conversation, followed by \textit{empathy} at 23.1\%, indicating that empathy plays an important role in such a psychiatric diagnosis-oriented dialogue.  

\begin{figure}[htbp]
    \centering
    \includegraphics[width=0.48\textwidth]{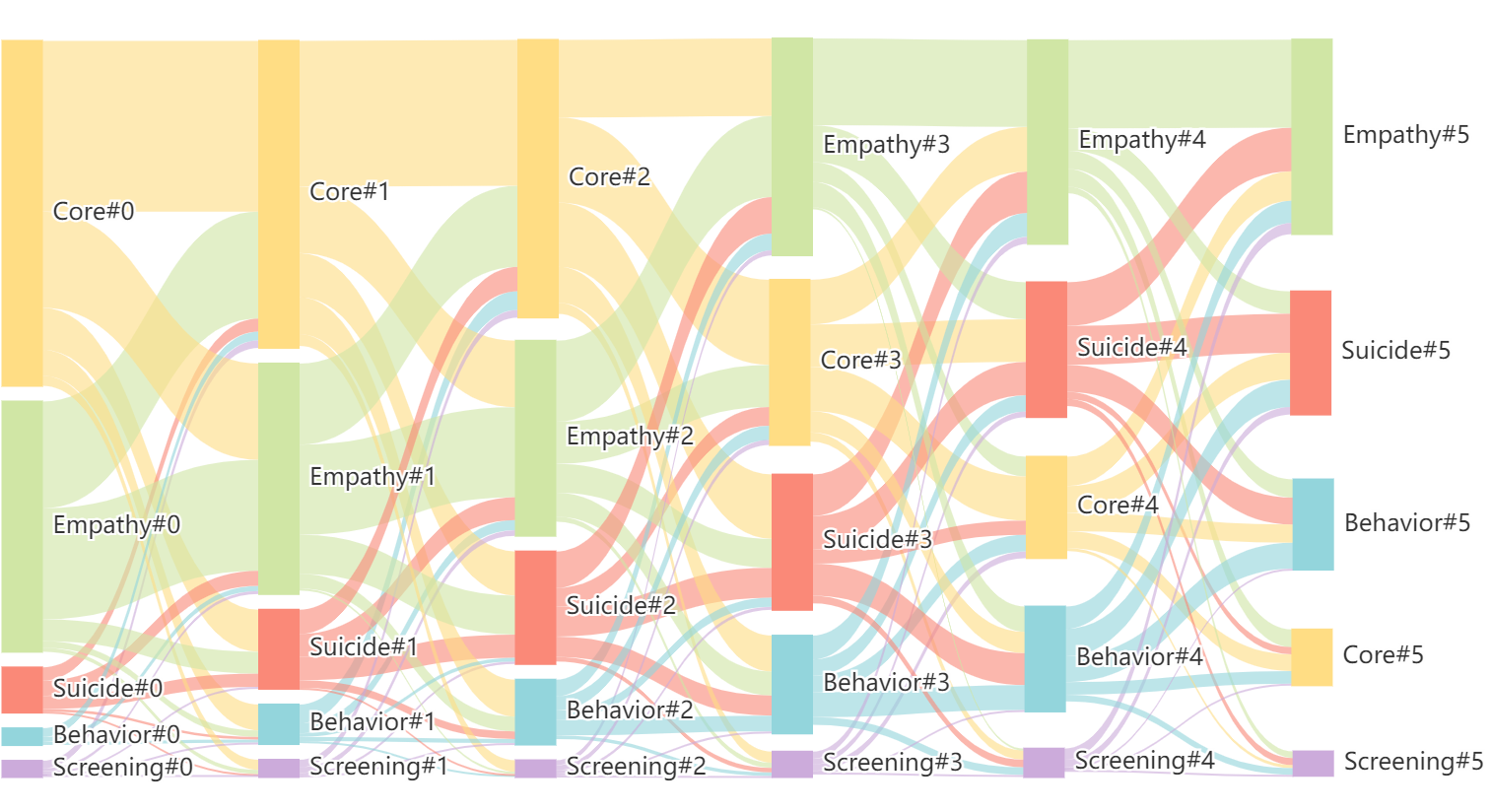}
    \caption{Topic Transitions. Topics over every 3 turns are visualized. The height represents the absolute number of dialogues at this topic. }
    \label{fig:Transition of topics}
\end{figure}

\paragraph{Topic Transition}
Figure \ref{fig:Transition of topics} illustrates the topic-transition process. Unlike other commonly seen dialogues where the topic rarely extends over one turn, diagnosis topics consistently occur across turns. Further, core symptoms like \textit{mood}, \textit{interest} are usually inquired in the beginning, gradually move to behavior symptom such as \textit{somatic symptom} and \textit{suicide}, which are normally experienced by severe patients. 
This echoes clinical practice where a consultation follows a gradual in-depth manner and provides emotional support from time to time.
\paragraph{Lexical Analysis of Empathy} 
As shown in Figure \ref{fig:topic proportion}, \textit{empathy} accounts for a large proportion, indicating its importance and commonness.
We extract its lexical features and observe that the empathy expressions in our dataset could mainly be divided into 4 aspects: 1) \texttt{understanding}: "will understand/is normal" to express understanding of the patient's situation; 2) \texttt{encouragement}: "is valuable" to help patients regain confidence; 3) \texttt{suggestions}: "you can try/try" to encourage patients to make changes and try; 4) \texttt{blessings}: "you will get well soon" to express blessings to the patient. 
In actual practice, providing empathetic and emotional support improves the medical experience and is a critical component of ensuring the success and completion of a diagnostic session\citep{hardy2019clinical}.
\section{Comparison with Related Datasets}
D$^4$ is compared with related datasets and manifested its characteristics as having more dialogue turns and utterances with a sufficient number of dialogues for model training (see Table \ref{tab: comparison with other dialogue dataset}). This again emphasizes that depression diagnosis is distinguished from current dialogue types and exhibits specific challenges with existing data.
\begin{table}[htbp]
    \centering
    \setlength\tabcolsep{3pt} 
    \resizebox{1\linewidth}{!}{
    \begin{tabular}{ccccc}
    \hline
    \textbf{Dataset}   &\textbf{Domain} & \textbf{Dialogues} & \textbf{Avg.turns} & \textbf{Avg.utterances} \\
    \hline
    MultiWOZ    & Restaurants, & \multirow{2}{*}{8,438}&\multirow{2}{*}{13.46}&\multirow{2}{*}{-}\\ 
    \citep{budzianowski2018multiwoz}&Hotels, etc&&&\\
    \hline
    MotiVAte &Mental  &\multirow{2}{*}{4,000}&\multirow{2}{*}{-}&\multirow{2}{*}{3.70}\\
    \citep{saha2021large}&Health     &&&\\
    \hline
    ESConv  &Emotional &\multirow{2}{*}{1,053}&\multirow{2}{*}{-}&\multirow{2}{*}{29.8}\\
    \citep{liu2021towards} &Support  &&&\\
    \hline
    MedDialog &Medical &\multirow{2}{*}{3,407,194}&\multirow{2}{*}{-}&\multirow{2}{*}{3.3}\\
    \citep{zeng2020meddialog} & Dialogue  &&&\\
    \hline
    DAIC-WOZ &Distress &\multirow{2}{*}{189}&\multirow{2}{*}{-}&\multirow{2}{*}{-} \\
    \citep{gratch2014distress} &Analysis  &&&\\
    \hline
    \multirow{2}{*}{\textbf{D$^4$(Ours)}}  &\textbf{Depression}  & \multirow{2}{*}{\textbf{1,339}}&\multirow{2}{*}{\textbf{21.55}}&\multirow{2}{*}{\textbf{60.91}}\\
     &\textbf{Diagosis}&&&\\
    \hline
    \end{tabular}
    }
    \caption{Comparison with Related Datasets}
    \label{tab: comparison with other dialogue dataset}
\end{table}
\paragraph*{Task-Oriented Dialogue Datasets}
Task-oriented dialogue dataset is one of the most essential components in dialogue systems study~\citep{ni2021recent}, consisting of various datasets for this purpose~\citep{chen2022dialogzoo}, i.e. MultiWOZ~\citep{budzianowski2018multiwoz}, MSR-E2E~\citep{li2018microsoft}, CamRest~\citep{wen2016conditional} , Frames~\citep{asri2017frames}. 
However, these dialogue datasets mainly involve daily scenarios instead of clinical practice.
Therefore, the number of dialogue turns is relatively small, with little attention paid to providing emotional support.  
\paragraph*{Emotional Support Datasets}
\par  A few dialogue studies on mental health address users' emotions in the dialogue process and endeavor to motivate users suffering from a mood disorder.
For example, \citet{saha2021large} presents the dialogue dataset MotiVAte to impart optimism, hope, and motivation for distressed people. Recently, works like ESConv~\citep{liu2021towards} switch their attention to construct a professional emotional support dialog Systems. However, they are mainly concerned with providing encouragement and advice to patients instead of providing professional diagnoses for screening purposes. 

\paragraph*{Medical Diagnosis Dialogue Datasets}
Some medical dialogue datasets target at diagnosis, such as MedDG~\citep{liu2020meddg} and MedDialog~\citep{zeng2020meddialog}. 
Meanwhile, some datasets aim at biomedical language
understanding such as CBLUE~\citep{zhang2021cblue}. However, these efforts focus mainly on somatic symptoms and physical diseases. MedDialog, although containing a small amount of psychiatric data, lacks professional psychiatric annotations, limiting its usage for a depression diagnosis dialogue system. It should be noted that the diagnosis process of depression essentially differs from that of somatic disorders. According to ICD-11~\citep{ICD11:2021}, in addition to somatic symptoms, patients often have multiple dimensions of symptoms such as mood, interest, mental status, and social function disorder. For this reason, psychiatrists need comprehensive information extracted from patients' subjective statements to provide unbiased diagnoses, leading to a longer, multi-domain dialogue process.
\paragraph*{Depression-Related Dialogue Dataset}
Along with the worldwide attention on depression, a few dialogue datasets strongly related to depression are constructed, such as DAIC-WOZ~\citep{gratch2014distress}, a multi-modal dataset. The dataset consists of face-to-face counseling conversations between a wizard interviewer and patients who suffer from depression, anxiety, etc. 
However, DAIC-WOZ only includes 189 dialogues without specific annotations, which is insufficient for dialogue generation training.

\section{Experiments}
\subsection{Tasks}
Upon the construction of D$^4$ with 1,339 well-annotated and up-to-clinical-standard depression diagnosis conversations, we can support an entire generation and diagnosis process mirroring the real-life clinical consultation scenario.
We split the entire depression diagnosis dialogue procedure into 4 subtasks: \textbf{Response Generation} aims to generate doctors' probable response based on the dialog context; \textbf{Topic Prediction} predicts the topic of the response based on the dialogue context. In our experiments, we jointly optimize the topic prediction model and the response generation model. We take the topic as a special first token of dialogue response; \textbf{Dialogue Summary} generates symptom summaries based on the entire dialog history; \textbf{Severity Classification} separately predicts the severity of depressive episodes and the suicide risk based on the dialogue context and dialogue summary. Binary (positive/negative) and fine-grained 4-class (positive further classed into mild, medium, and severe) classifications are both investigated. 
\subsection{Backbone Models}
We use Transformer~\citep{vaswani2017attention} pretrained on MedDialog~\citep{zeng2020meddialog}, BART~\citep{lewis2019bart} pretrained on Chinese datasets~\citep{shao2021cpt}, CPT~\citep{shao2021cpt} and BERT~\citep{devlin2019bert} as  backbone models to conduct the experiments.
\subsection{Objective Evaluation}
\paragraph{Generation and Summarization}
We evaluate the \emph{response generation task} and \emph{dialog summary task} with objective metrics including BLEU-2~\citep{papineni2002bleu}, Rouge-L~\citep{lin2004rouge}, METEOR~\citep{banerjee2005meteor} to measure the similarity between model generated responses and labels. To show the generation diversity, we also compute DIST-2~\citep{li2015diversity}.
We implement jieba\footnote{https://github.com/fxsjy/jieba} for tokenization and compute the metrics at the word level. 
\par Results for the \textit{response generation task} are presented in Table \ref{tab: exp_tod_result}. Five observations can be drawn: 1) BART and CPT exhibit similar generation performance on our dataset; 
2) Both models vastly outperform Transformer, which is pretrained on the medical corpus, suggesting that, on the one hand, pertrained language models with more parameters could improve generation performance; on the other hand, depression diagnosis is different from traditional somatic-oriented medical dialogues;
3) Based on the topic of response predicted by the model itself, the model could generate a more accurate response, which is of great significance for the model to be applied in real human-machine interaction scenarios; 4)Based more accurate topics predicted by BERT, response generation performance is enhanced, indicating that higher topic prediction accuracy can effectively improve generation accuracy.
5) Given golden topics, generation performance can be further enhanced. 

\par \textit{Topic Prediction} accuracy results are shown as Topic ACC. in Table \ref{tab: exp_tod_result}. We adopted the topic category regrouped in \ref{fig:topic proportion} and similar trend is observed: BART $\approx$ CPT $\textgreater$ Transformer. The F1 of each topic (see Table \ref{tab: F1 of Topics Predicted by BART}) shows that the accuracy of empathy is the bottleneck of this task, indicating that proper timing of empathy remains challenging for models and is a potential direction for further work.    
\begin{table}[]
    \centering
    \resizebox{1\linewidth}{!}{
    \begin{threeparttable}
    \begin{tabular}{cccccc}
    \hline
    \textbf{Model}   & \textbf{BLEU-2} & \textbf{ROUGE-L} & \textbf{METEOR} & \textbf{DIST-2}&\textbf{Topic ACC.}  \\
    \hline
    Transformer-    &7.28\%    &0.21   & 0.1570    & 0.29 &- \\
    BART-              &19.29\%        &0.35   & 0.2866        & 0.09 &-\\
    CPT-                &19.79\%        &0.36   & 0.2969        &0.07 &- \\
    \hline
    Transformer & 13.43\% & 0.33 & 0.2620 & 0.04 & 36.82\% \\
    BART & 28.62\% & 0.48 & 0.4053 & 0.07 & 59.56\%\\
    CPT & 29.40\% & 0.48 & 0.4142 & 0.06 & 59.77\%\\
    \hline
    Transformer-BERT & 23.95\% & 0.40 & 0.3758 &0.22&61.32\%\\
    BART-BERT & 33.73\% & 0.50 & 0.4598 & 0.07&61.32\% \\
    CPT-BERT & 34.64\% & 0.51 & 0.4671 & 0.06 & 61.32\%\\
    \hline
    Transformer* & 25.37\% & 0.41 & 0.3905 & 0.04 & - \\
    BART* & 37.02\% & 0.54 & 0.4920 & 0.07 & - \\
    \textbf{CPT*} & \textbf{37.45\%} & \textbf{0.54} & \textbf{0.4943} & \textbf{0.06} & \textbf{-}\\
    \hline
    \end{tabular}
    \begin{tablenotes}
        \item $-$ means topics are excluded, BERT means topics predicted by BERT are given as prompt, $*$ means golden topics are given as prompt
      \end{tablenotes}
    \end{threeparttable}}
    
    \caption{Evaluation Results of Response Generation and Topic Prediction}
    \label{tab: exp_tod_result}
\end{table}
\begin{table}[]
    \centering
    \resizebox{1\linewidth}{!}{
    \begin{threeparttable}
    \begin{tabular}{cccccc}
    \hline
    &\textbf{Core}& \textbf{Behavior} & \textbf{Empathy} & \textbf{Suicide} &\textbf{Screening} \\
    \hline
    \textbf{F1}&0.63&0.69&0.24&0.49&0.35\\
    \hline
    \end{tabular}
    \end{threeparttable}
    }
    \caption{F1 of Topics Predicted by BART}
    \label{tab: F1 of Topics Predicted by BART}
\end{table}
\par Results for \textit{Dialog Summary} are listed in Table \ref{tab: summary metrics}, CPT is on par with BART regarding the N-gram overlap with human references. Nevertheless, CPT exhibits a higher DIST-2 score, suggesting its superiority in generation diversity. We manually annotated summaries by 13 symptoms from ICD-11 and calculated the summaries' sample average F1 score on the multi-label classification task of symptoms, where CPT and BART perform the same.
It shows that the summary generated by the model can accurately summarize most symptoms.
 
\begin{table}[htbp]
    \centering
    \resizebox{1\linewidth}{!}{
    \begin{tabular}{cccccc}
    \hline
    \textbf{Model}     &  \textbf{BLEU-2} & \textbf{ROUGE-L} & \textbf{METEOR} & \textbf{DIST-2}  &\textbf{Symptom F1}\\
    \hline
BART               &16.44\%        &0.26   & 0.25      & 0.19 & 0.67 \\
\textbf{CPT}        &\textbf{16.45\%}       &\textbf{0.26}   & \textbf{0.24}        & \textbf{0.21}& \textbf{0.68} \\
    \hline
    \end{tabular}}
    \caption{Evaluation Results of Dialog Summary Task}
    \label{tab: summary metrics}
\end{table}

\paragraph{Severity Classification} Binary and 4-class classification are evaluated by average weighted precision, recall, and F1 by sklearn\footnote{https://scikit-learn.org}, and results of depression severity and suicide risk severity are shown in Table \ref{tab: Depression Severity Classification Results} and Table \ref{tab: Suicide Severity Classification Results}. For the classification of depression severity, we conducted experiments based on dialogue history and symptom summaries respectively. The evaluation results show that the accuracy of 2-classification and 4-classification based on summaries is significantly improved, indicating that symptom summaries have extracted vital information from the dialogue, being extremely helpful for diagnosis. Although the results of 4-classification tasks are relatively poor compared with the performance in 2-classification tasks, as a screening tool, the binary classification results are already sufficient in the practical application of the system.

\begin{table}[htbp]
    \centering
    \resizebox{1\linewidth}{!}{
    \begin{tabular}{cccccc}
    \hline
    \textbf{Task} &\textbf{Input}& \textbf{Model} &  \textbf{Precision} &\textbf{Recall} & \textbf{F1}  \\
    \hline
    \multirow{6}{*}{2-class}&\multirow{3}{*}{dialog} & BERT & 0.81 \footnotesize$\pm.04$ & 0.80\footnotesize$\pm.03$ & 0.80 \footnotesize$\pm.03$ \\
          & & BART & 0.80\footnotesize$\pm.02$ & 0.79\footnotesize$\pm.03$ & 0.79\footnotesize$\pm.03$\\
        & & CPT  & 0.79\footnotesize$\pm.02$ & 0.78\footnotesize$\pm.03$ & 0.78\footnotesize$\pm.03$  \\
    \cline{2-6}
    & \multirow{3}{*}{summary}& BERT &0.90\footnotesize$\pm.02$&0.90\footnotesize$\pm.02$&0.90\footnotesize$\pm.02$\\
    & & BART &0.89\footnotesize$\pm.03$&0.89\footnotesize$\pm.03$&0.89\footnotesize$\pm.03$\\
    & & \textbf{CPT} &\textbf{0.92\footnotesize$\pm.01$}&\textbf{0.92\footnotesize$\pm.02$}&\textbf{0.92\footnotesize$\pm.01$}\\
    \hline
    \multirow{6}{*}{4-class}                    
          & \multirow{3}{*}{dialog}  & BERT & 0.49\footnotesize$\pm.05$ & 0.45\footnotesize$\pm.04$ & 0.45\footnotesize$\pm.04$ \\
           & & BART & 0.53\footnotesize$\pm.04$ & 0.53\footnotesize$\pm.04$ & 0.52\footnotesize$\pm.04$\\
           & & CPT  & 0.49\footnotesize$\pm.04$  & 0.47\footnotesize$\pm.04$  & 0.46\footnotesize$\pm.05$ \\
    \cline{2-6}    
    & \multirow{3}{*}{summary}& BERT&0.67\footnotesize$\pm.04$&0.66\footnotesize$\pm.04$&0.66\footnotesize$\pm.04$\\
    & & BART &0.68\footnotesize$\pm.03$&0.67\footnotesize$\pm.02$&0.66\footnotesize$\pm.02$\\
    & & \textbf{CPT} &\textbf{0.73\footnotesize$\pm.03$}&\textbf{0.72\footnotesize$\pm.03$}&\textbf{0.72\footnotesize$\pm.03$}\\
    \hline
    \end{tabular}}
    \caption{Depression Severity Classification Results}
    \label{tab: Depression Severity Classification Results}
\end{table}
\begin{table}[htbp]
    \centering
    \resizebox{1\linewidth}{!}{
    \begin{tabular}{cccccc}
    \hline
    \textbf{Task} &\textbf{Input}& \textbf{Model} &  \textbf{Precision} &\textbf{Recall} & \textbf{F1}  \\
    \hline
    \multirow{3}{*}{2-class}&\multirow{3}{*}{dialog} & BERT & 0.81\footnotesize$\pm.02$ & 0.78\footnotesize$\pm.02$ & 0.79\footnotesize$\pm.02$ \\
          & & BART & 0.77\footnotesize$\pm.02$ & 0.75\footnotesize$\pm.02$ & 0.75\footnotesize$\pm.02$\\
        & & \textbf{CPT}  & \textbf{0.84\footnotesize$\pm.02$} & \textbf{0.82\footnotesize$\pm.03$} & \textbf{0.82\footnotesize$\pm.03$} \\
    \hline
    \multirow{3}{*}{4-class}                    
          & \multirow{3}{*}{dialog}  & BERT & 0.72\footnotesize$\pm.03$ & 0.64\footnotesize$\pm.04$ & 0.66\footnotesize$\pm.03$ \\
           &  & BART & 0.70\footnotesize$\pm.05$ & 0.66\footnotesize$\pm.04$ & 0.65 \footnotesize$\pm.03$\\
           & & \textbf{CPT}  & \textbf{0.76\footnotesize$\pm.02$} & \textbf{0.68\footnotesize$\pm.02$}& \textbf{0.70\footnotesize$\pm.02$} \\
    \hline
    \end{tabular}}
        \caption{Suicide Severity Classification Results}
    \label{tab: Suicide Severity Classification Results}
\end{table}
\subsection{Human Interactive Evaluation}
To comprehensively evaluate the model's conversation experience with the user, we include human interactive evaluation for CPT  with a rule-based chatbot.
Evaluators were invited to chat with both bots in a random order upon the provided patient portrait and rated on 4 aspects with a 1-5 scale: \textbf{Fluency} measures how fluently the conversation flows;  \textbf{Comforting} measures how comforting the responses are; \textbf{Doctor-likeness} measures to what extent does the chatbot flexibly adjust the topic according to the patient's description; \textbf{Engagingness} measures to what time could the chatbot maintain their attention to continue the chat.
\begin{table}[htbp]
    \centering
    \resizebox{1\linewidth}{!}{
    \begin{threeparttable}
    \begin{tabular}{ccccc}
    \hline
    \multirow{1}{*}{\textbf{CPT vs Base}} & \multicolumn{4}{c}{\textbf{Metric}} \\
    \textbf{Result}&\textbf{Fluency}& \textbf{Comforting} &  \textbf{Doctor-likeness} &\textbf{Engagingness}  \\
    \hline
    
    \textbf{Win}&\textbf{19$^{\dagger}$}&\textbf{18$^{\dagger}$}&\textbf{17$^{\dagger}$}&\textbf{20$^{\dagger}$}\\
    \textbf{Lose}&14&13&14&11\\
    \hline
    \end{tabular}
     \begin{tablenotes}
        \item $\ddagger$/$\dagger$means p-value $<$ 0.05/0.5 respectively
      \end{tablenotes}
    \end{threeparttable}
    }
    \caption{Human Evaluation Results}
    \label{tab: Human Evaluation Results}
\end{table}
\par The interactive human evaluation results are illustrated in Table \ref{tab: Human Evaluation Results}. 
The CPT-based chatbot outperforms rule-based bots on all four metrics, suggesting that dialogue models can help us build more human-like and user-friendly depression diagnosis systems. 
In particular, the discrepancy in engagingness indicates that users prefer chatbots that can better understand and comfort users in completing the depression screening process. We give some empathy examples of human interactive evaluation in Table~\ref{tab: Empathy Words of Human Evaluation Results}, indicating that the model can generate diverse empathy representations from different aspects.
\begin{table}[hbtp]
    \centering
    \resizebox{1\linewidth}{!}{
    \begin{threeparttable}
    \begin{tabular}{cc}
    \hline
    \textbf{Aspects}&\textbf{Examples} \\
    \hline
    \multirow{2}{*}{Understanding}& \textit{I could understand you.}\\
    &\textit{I could understand your feelings.}\\
    \hline
    \multirow{2}{*}{Encouragement}& \textit{Everyone has their own value. }\\
    & \textit{Everyone has their own characteristics.}\\
    \hline
    \multirow{2}{*}{Suggestion}&\textit{It is suggested to seek professional }\\
    &\textit{medical help as soon as possible.}\\
    \hline
    \multirow{2}{*}{Blessing}&\textit{Wish you a happy life!}\\
    &\textit{Hope you get well soon!}\\
    \hline
    \end{tabular}
    \end{threeparttable}
    }
    \caption{Empathy Examples in Human Evaluation}
    \label{tab: Empathy Words of Human Evaluation Results}
\end{table}

\section{Conclusion}
In this paper, we designed a 3-phase data collection and constructed a close-to-clinical-practice and up-to-clinical-standard depression diagnosis dataset with 1,339 conversations accompanied by psychiatrists' diagnosis summaries. Further, we conduct experimental validation on multiple tasks with state-of-art models and compare the results with objective and human evaluation. 
The evaluation results show that the model-based chatbot outperforms traditional rule-based dialogue bots in all metrics, indicating that a more user-friendly dialogue system can be built with our dataset. However, the model is still not effective enough in generating appropriate empathic responses suggesting that the model needs further improvement to generate more appropriate empathy during the consultation process.

\section*{Limitations}
Our work has some limitations. The principal limit of our work is that our dataset D$^4$ is in Chinese, which in line with Chinese culture and expression habits. Therefore, it may not be applicable to translate the conversations into another language directly, so further exploration is required for our work to transfer to other languages. However, considering that there are no similar datasets in other languages published before, 
we hope that our data collection method and data form (dialog+summary+diagnosis) could inspire more research on this unique type of dialogue in the future.
\par Additionally, for patient privacy protection, our dialogue data is collected in a simulated manner, not from real scenarios. This approach helps construct a more secure and generalizable consultation dialogue system because we have defined the acting doctors' behaviors during the data collection process, that is, the system behavior range. But it should be mentioned that our dataset cannot restore the expressions of actual patients and doctors. For this reason, the textual features of acting patients in our dataset are not sufficient for the classification of depression. Therefore, it is meaningful to explore the construction of a more empathy-driven and diagnostic-accurate consultation dialogue system based on our dataset rather than conduct textual depression classification.

\section*{Ethics Statement}
This research study has been approved by the Ethics Review Board at the researchers’ institution (Ethics Approval No. I2022158P). Different stages in data collection comply with corresponding ethical requirements and we endeavour to protect privacy and respect willingness of our data providers and annotators. 

Specifically, our data collection falls under the Personal Information Protection Law of the People's Republic of China. In the phase of portrait collection, the collection application was developed as a WeChat mini program\footnote{https://developers.weixin.qq.com/miniprogram/en/dev}, which complied with the privacy protection agreement and passed the security and privacy check of WeChat mini program before releasing on the platform. Furthermore, all the portrait providers signed an informed consent form to give permission to collect their anonymous information for research purposes. 

In the phase of the dialogue collection process, all the workers and annotators are informed about the purpose of our data collection and equally paid for their workload. In the phase of the dialogue examination process, the psychiatrists and psychotherapists are licensed to practice and paid equally for their workload.

To protect users' privacy, we anonymized the portraits by storing them without a one-to-one correspondence between the identification information required for user login and the data we use in research. Therefore, all the information that could uniquely identify individual people is excluded from our dataset and research process. Regarding offensive content, we rigorously filtered the dataset manually to ensure that it did not contain any offensive content or words encouraging patients to self-harm and commit suicide.
We will also require the users of D$^4$ to comply with a data usage agreement to prevent the invasion of privacy or other potential misuses.


\section*{Acknowledgements}
This project is supported by National Natural Science Foundation of China (Grant No.61901265, 92048205), Shanghai Municipal Science and Technology Major Project (2021SHZDZX0102), SJTU Medicine-Engineering Project (No.YG2020YQ25), and Xuhui District Artificial Intelligence Medical Hospital Cooperation Project (2021-005).

\bibliography{anthology,custom}
\bibliographystyle{acl_natbib}
\newpage
\appendix

\section{Data Example}
\label{sec: Data Example}
\begin{figure}[htbp]
    \includegraphics[width=0.48\textwidth]{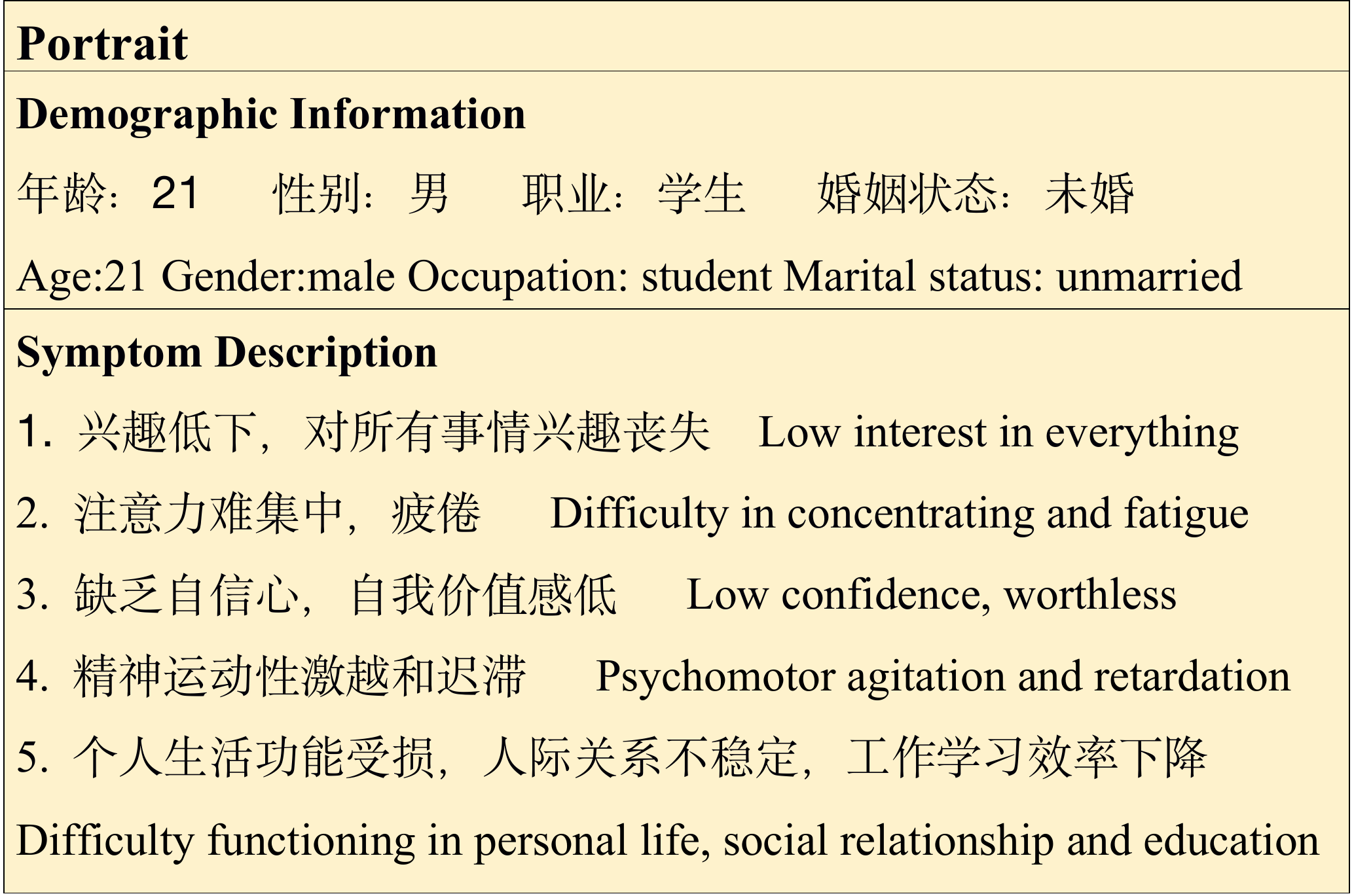}
    \caption{A Portrait Example}
    \label{fig: Potrait Example}
\end{figure}

\begin{figure}[htbp]
    \includegraphics[width=0.48\textwidth]{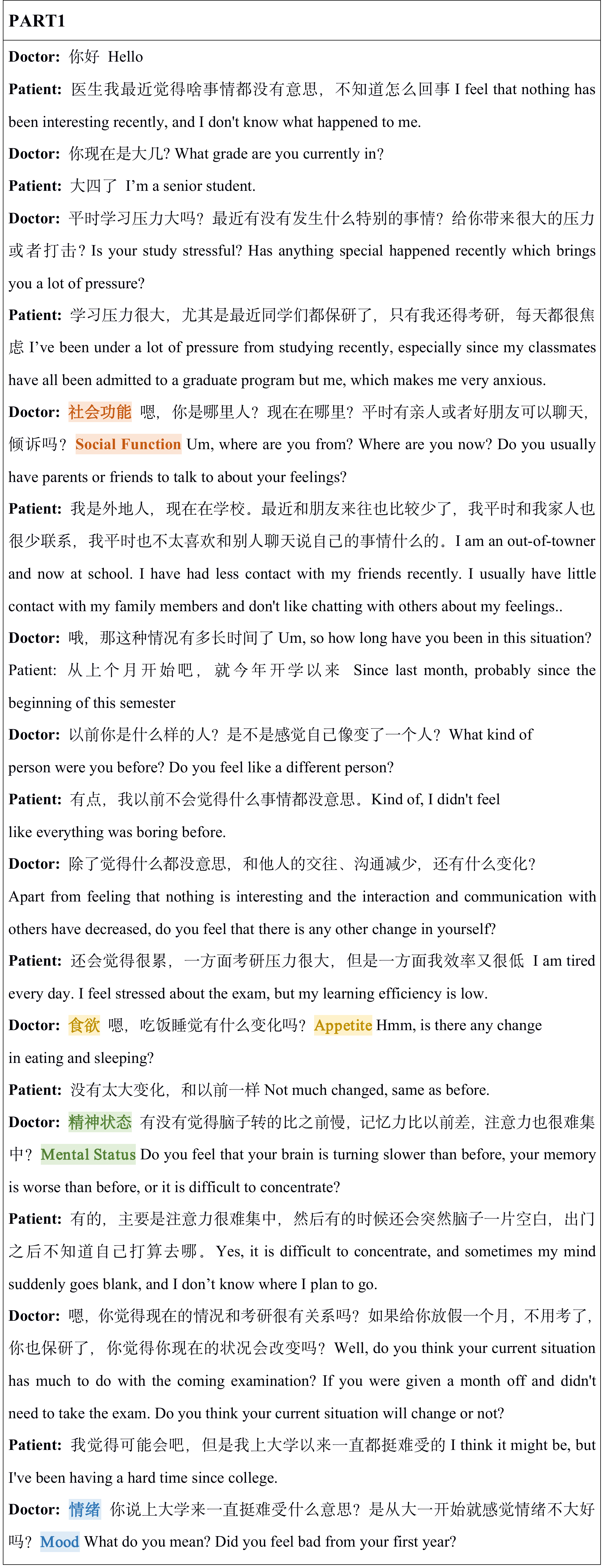}
    \caption{A Dialogue Example - Part1}
    \label{fig: A Dialogue Example - Part1}
\end{figure}

\begin{figure}[htbp]
    \centering
    \includegraphics[width=0.48\textwidth]{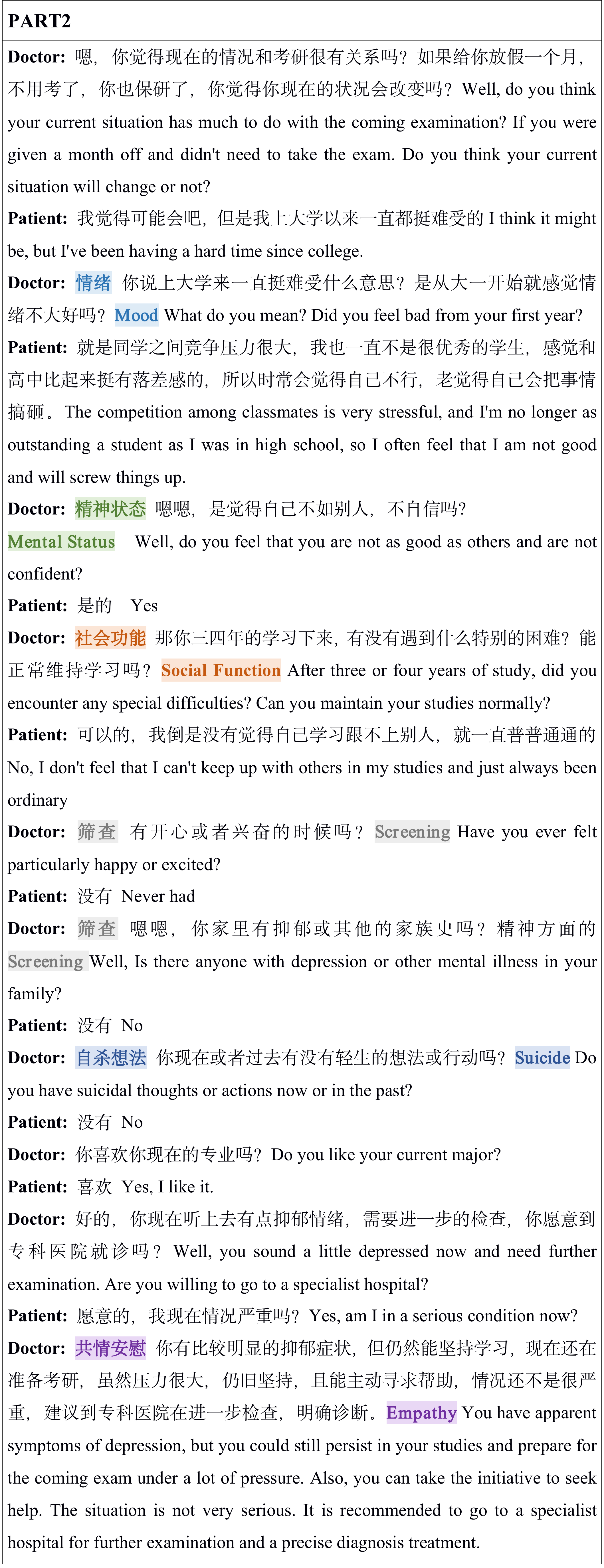}
    \caption{A Dialogue Example - Part2}
    \label{fig: A Dialogue Example - Part2}
\end{figure}

\begin{figure}
    \includegraphics[width=0.48\textwidth]{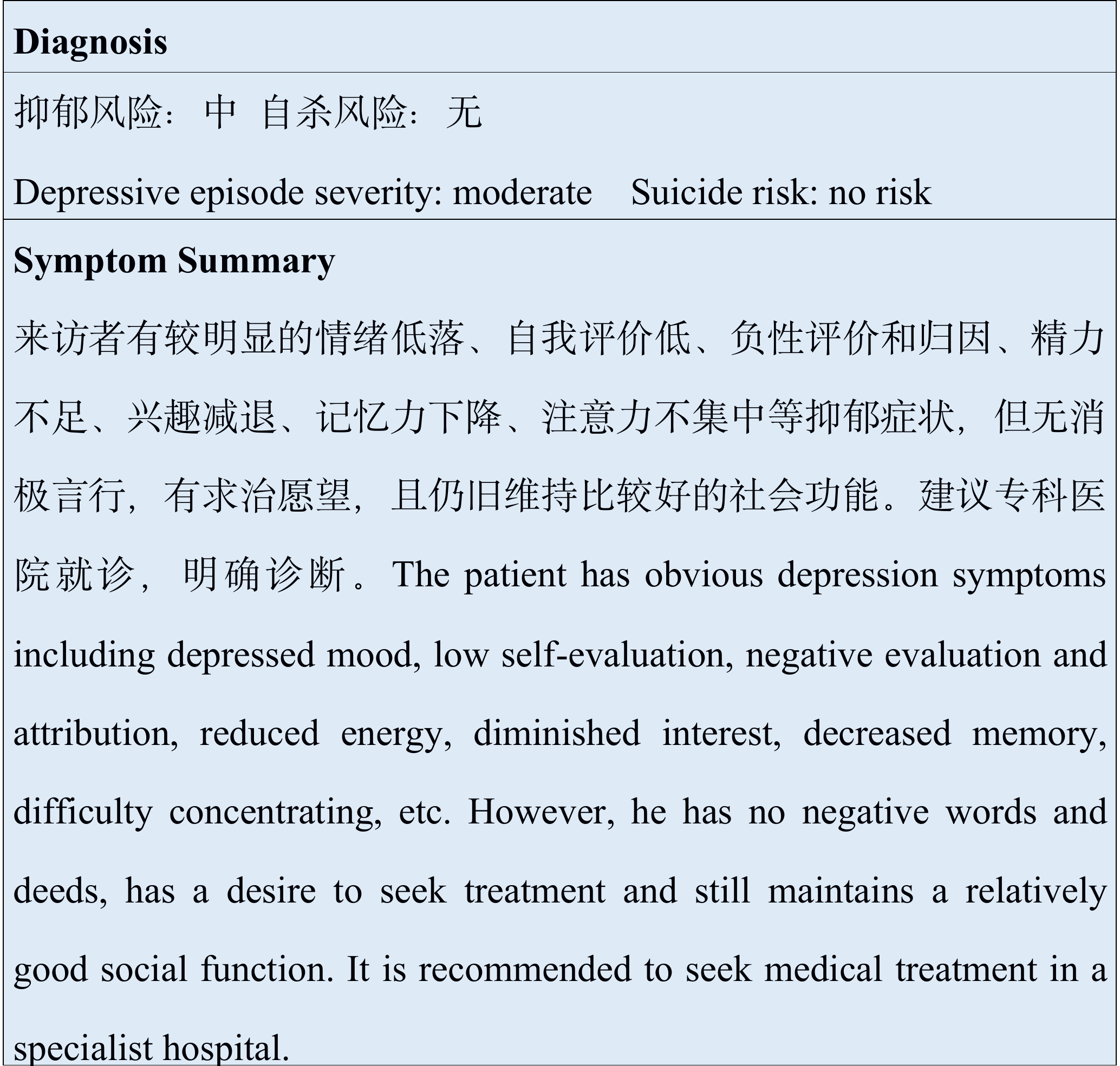}
    \caption{A Diagnosis Example}
    \label{fig: Diagnosis Example}
\end{figure}

The portrait (Figure \ref{fig: Potrait Example}), the dialogue (Figure \ref{fig: A Dialogue Example - Part1} and Figure \ref{fig: A Dialogue Example - Part2}), and diagnosis (Figure \ref{fig: Diagnosis Example}) belong to the same data example in our dataset. We marked the topic (if any) of the doctor's responses in the conversation history. In this example, the doctor combined sleep and appetite into one question, so only one topic of appetite was marked. In addition, for the convenience of presentation, we have combined the doctor's multiple utterances of the same turn into one sentence. 
To compare machine generation performance with humans, we provide data examples of the same portrait in this section and Section \ref{ssec:Human Interactive Example} - Human Interactive Example. Dialogues in D$^4$ were simulated based on diverse portraits showed in Section \ref{ssection: Statistics of Portraits' Demographic Information}. More data examples can be found in website \href{https://x-lance.github.io}https://x-lance.github.io/D4. 
\section{Data Characteristics}
\begin{figure}[htbp]
    \includegraphics[width=0.48\textwidth]{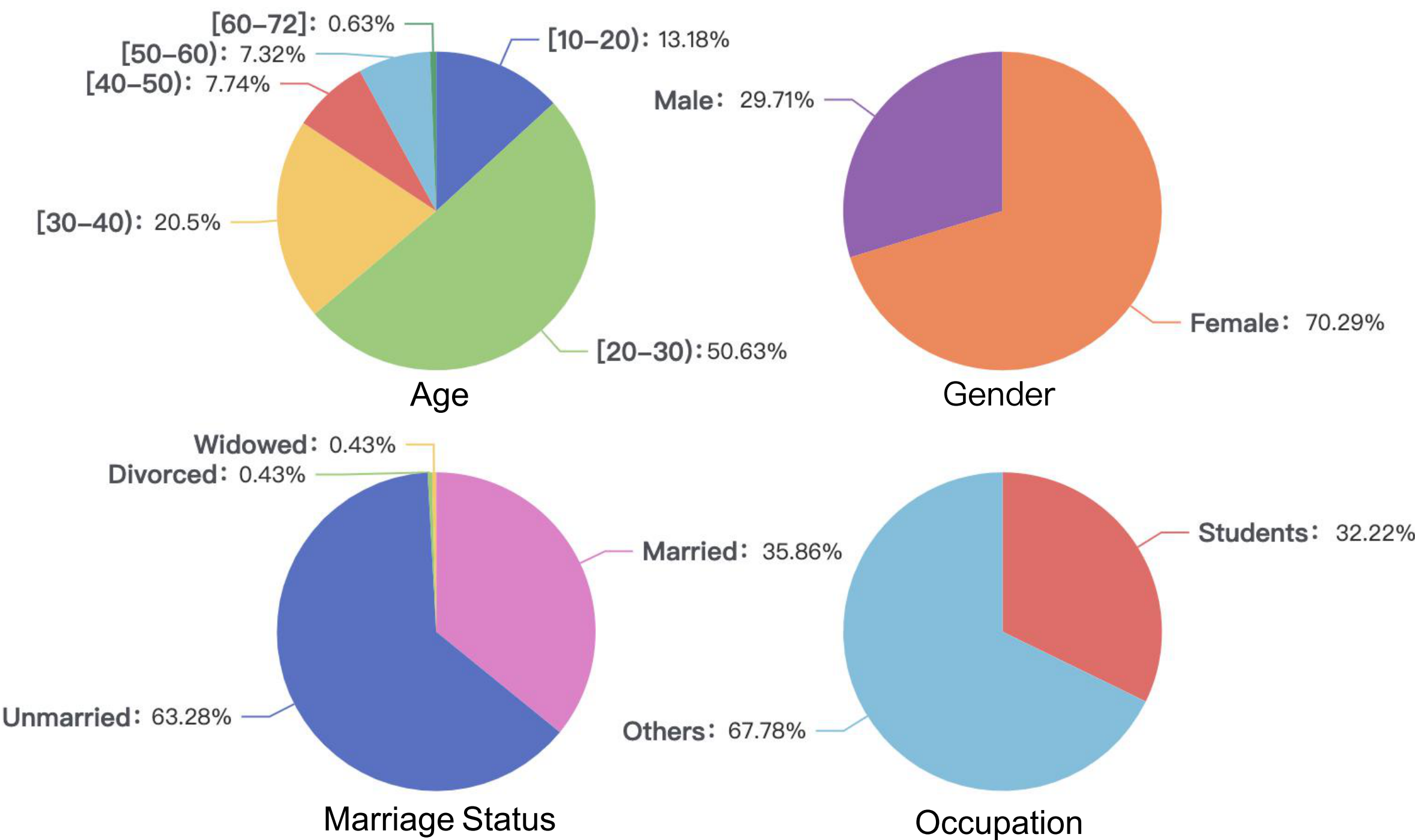}
    \caption{Aggregated Demographic Information}
    \label{fig: Aggregated Potraits' Demographic Information}
\end{figure}

\begin{figure}[htbp]
    \centering
    \includegraphics[width=0.48\textwidth]{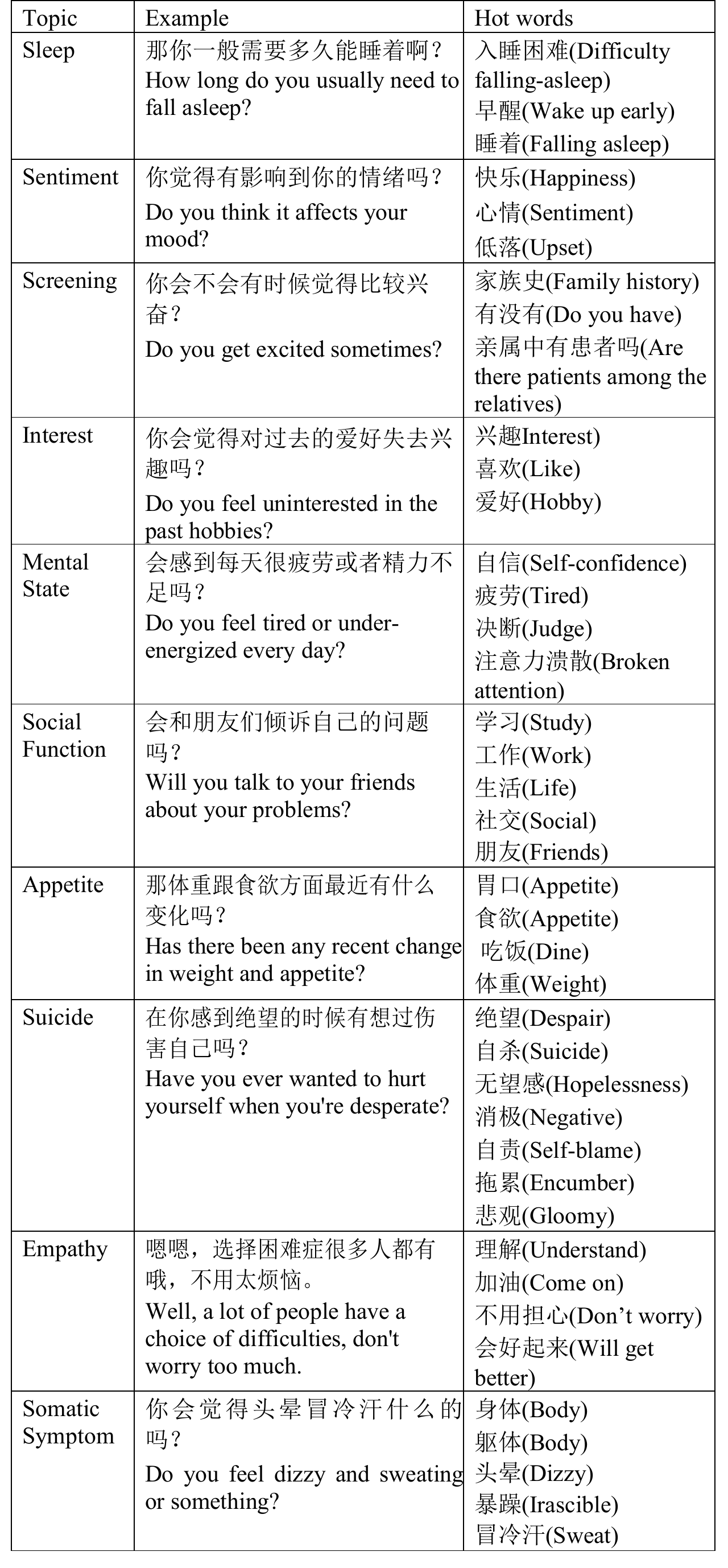}
    \caption{Samples of Doctors' Topic}
    \label{fig: Samples of doctors' topic}
\end{figure}

\begin{figure}[htbp]
    \centering
    \includegraphics[width=0.35\textwidth]{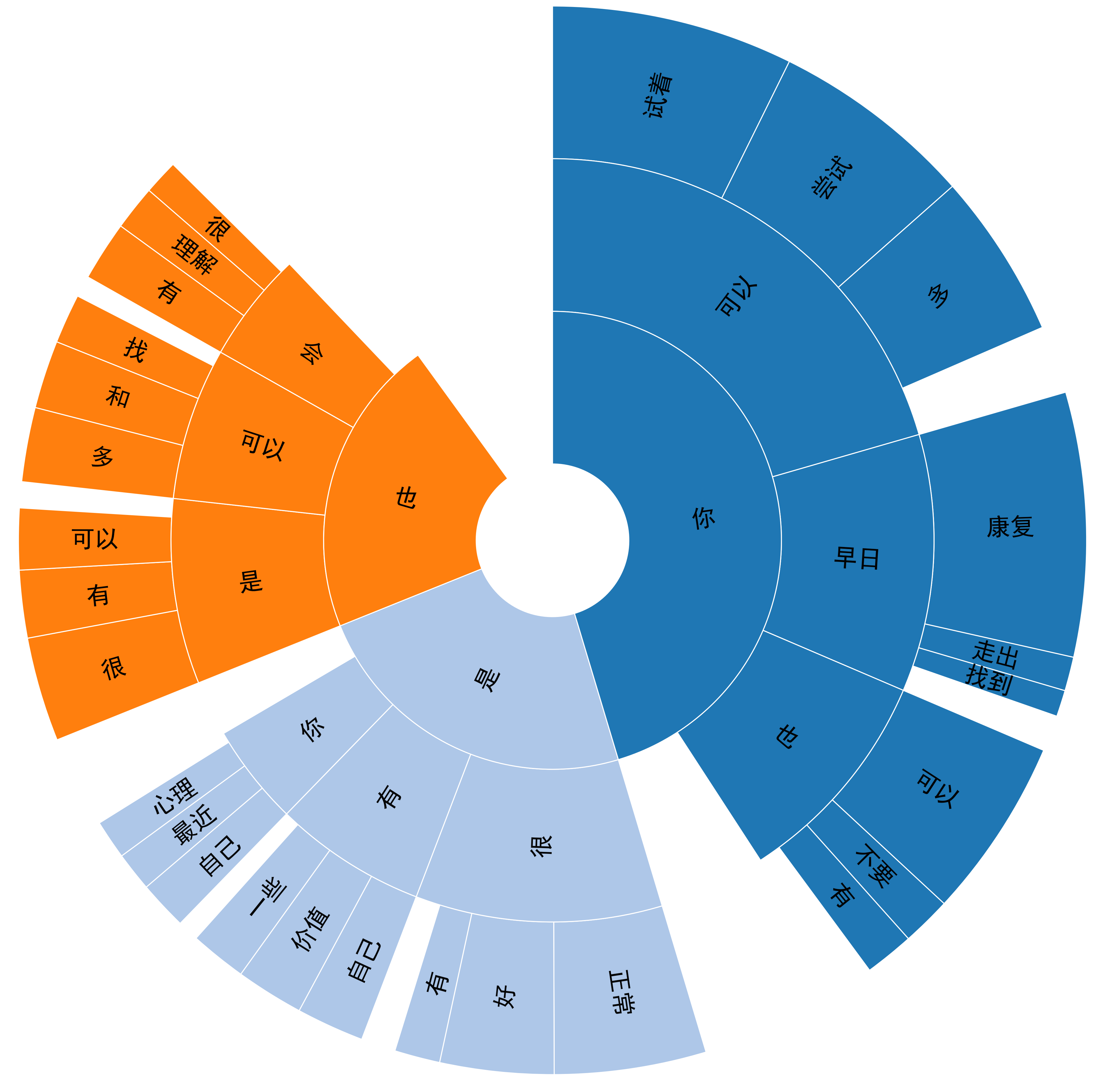}
    \caption{Lexical Feature of Empathy}
    \label{fig:Lexical Feature of Empathy}
\end{figure}

\paragraph{Statistics of Portraits' Demographic Information}
\label{ssection: Statistics of Portraits' Demographic Information}
The aggregated demographic information of 478 portraits is provide in Figure \ref{fig: Aggregated Potraits' Demographic Information}.
\paragraph{Topic Examples} In Figure \ref{fig: Samples of doctors' topic}, we present the 10 topics with their typical examples and hot words.
\paragraph{Lexical Feature of Empathy}
In Figure \ref{fig:Lexical Feature of Empathy}, we show the lexical feature of empathy words in our dataset in the sunburst figure.
\section{Backbone Model Introduction}
\paragraph{Rule-based Model} Without existing chatbots having the same function, we built the rule-based chatbot by state machine as the baseline. Based on ICD-11~\citep{ICD11:2021}, DSM-5~\citep{APA2013}, the bot covers the same topics as the dialogue simulation process mentioned in 2.2.2. This robot has fixed question templates and recognizes the user's answer based on regular matching, based on which it performs state jumps until all symptom information is acquired. 
\paragraph{Transformer} We use the classic sequence-to-sequence model~\citep{vaswani2017attention} to conduct the response generation and topic prediction experiment. The implementation used is HuggingFace\footnote{https://github.com/huggingface/transformers}. The parameters are loaded from the transformer pretrained on MedDialog~\citep{zeng2020meddialog}, a Chinese Medical Dialogue Dataset.

\paragraph{BART} BART~\citep{lewis2019bart} is a denoising sequence-to-sequence pre-trained model, which is a start-of-art model for both text generation and summary tasks. For this reason, we use Bart pretrained on Chinese datasets~\citep{shao2021cpt} to conduct the response generation and dialog summary task.
\paragraph{CPT} CPT~\citep{shao2021cpt} is a novel Chinese pre-trained un-balanced transformer model, which is not only effective in generation tasks but also has powerful classification ability, so we choose it as our backbone model to conduct the generation task and also compare its performance of classification task with BART.
\paragraph{BERT} Bert \citep{devlin2019bert} is effectively used for a wide range of language understanding tasks, such as question answering and language inference. Thus, we use the version\footnote{https://huggingface.co/hfl/chinese-macbert-base} which is pre-trained on eight popular Chinese NLP datasets, to conduct the classification task.
\section{Training Details}
The division of train, validation, and test sets for all experiments is close to 8:1:1, and the data of different depression severity are also internally distributed according to the above ratio.
\paragraph{Response Generation}
For BART and CPT models, the initial parameters are pretrained on Chinese datasets~\citep{shao2021cpt}. We use a cosine learning rate scheduler with the initial learning rate of 1e-5, 100 warm-up steps, and the AdamW optimizer~\citep{DBLP:conf/iclr/LoshchilovH19}. Beam search where the number of beams is 4 is used in response generation. Models are trained for 30 epochs. The one with the best BLEU-2 metric on the evaluation set is selected for the test.

For the Transformer, we use the implementation by HuggingFace\footnote{https://github.com/huggingface/transformers}. We load the parameters of the Transformer pretrained on MedDialog~\citep{zeng2020meddialog}. The weight parameters were learned with Adam and a linear learning rate scheduler with the initial learning rate of 1.0e-4 and 100 warm-up steps. The batch size was set to 16. Top-$k$ random sampling ~\citep{fan2018hierarchical} is used in response generation. The model is trained for 20 epochs. The one with the highest BLEU-2 score on the evaluation set is chosen for the test.

We spliced multiple sentences of the doctor in the same round into the dialogue history, and selected the last topic as the topic of the new sentence. Due to the limitation of models' positional embedding, we intercepted data with a length over 512. In the response generation task, we try to keep the most recent conversations as they are more instructive to the current response. 

\paragraph{Dialog Summary}
Both BART and CPT models are trained for 50 epochs. We use a cosine learning rate scheduler with the initial learning rate of 1e-5 and 100 warm-up steps and the AdamW optimizer. The one with the highest rouge-1 metric on the evaluation set is selected for the test.

If the input dialog history is longer than the model's input size, we retain the 512 tokens in the middle of the dialog.
\paragraph{Severity Classification}
For BERT, BART, and CPT models, we use a cosine learning rate scheduler with the initial learning rate of 1e-5, 100 warm-up steps, and the AdamW
optimizer~\citep{DBLP:conf/iclr/LoshchilovH19}. Models are trained for 30 epochs. The one with the best F1-score metric on the evaluation set is selected for the test.

For the classification based on dialog history, we retain 512 tokens in the middle of the dialog. For the classification based on dialog summary, we retain 128 tokens in the middle of the summary.
\section{Generation Examples }
\begin{figure}[htbp]
    \centering
    \includegraphics[width=0.48\textwidth]{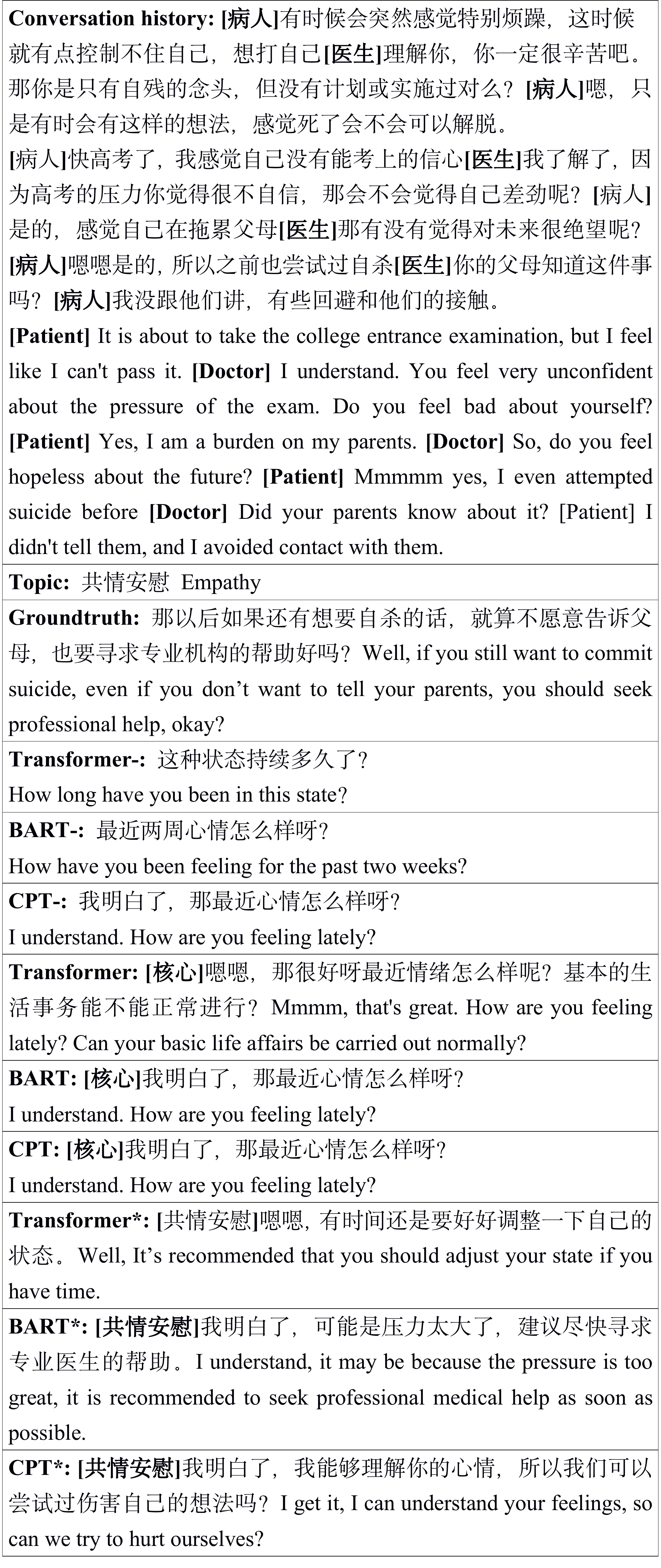}
    \caption{Examples of generated response}
    \label{fig: Examples of generated response}
\end{figure}

\begin{figure}[htbp]
    \centering
    \includegraphics[width=0.48\textwidth]{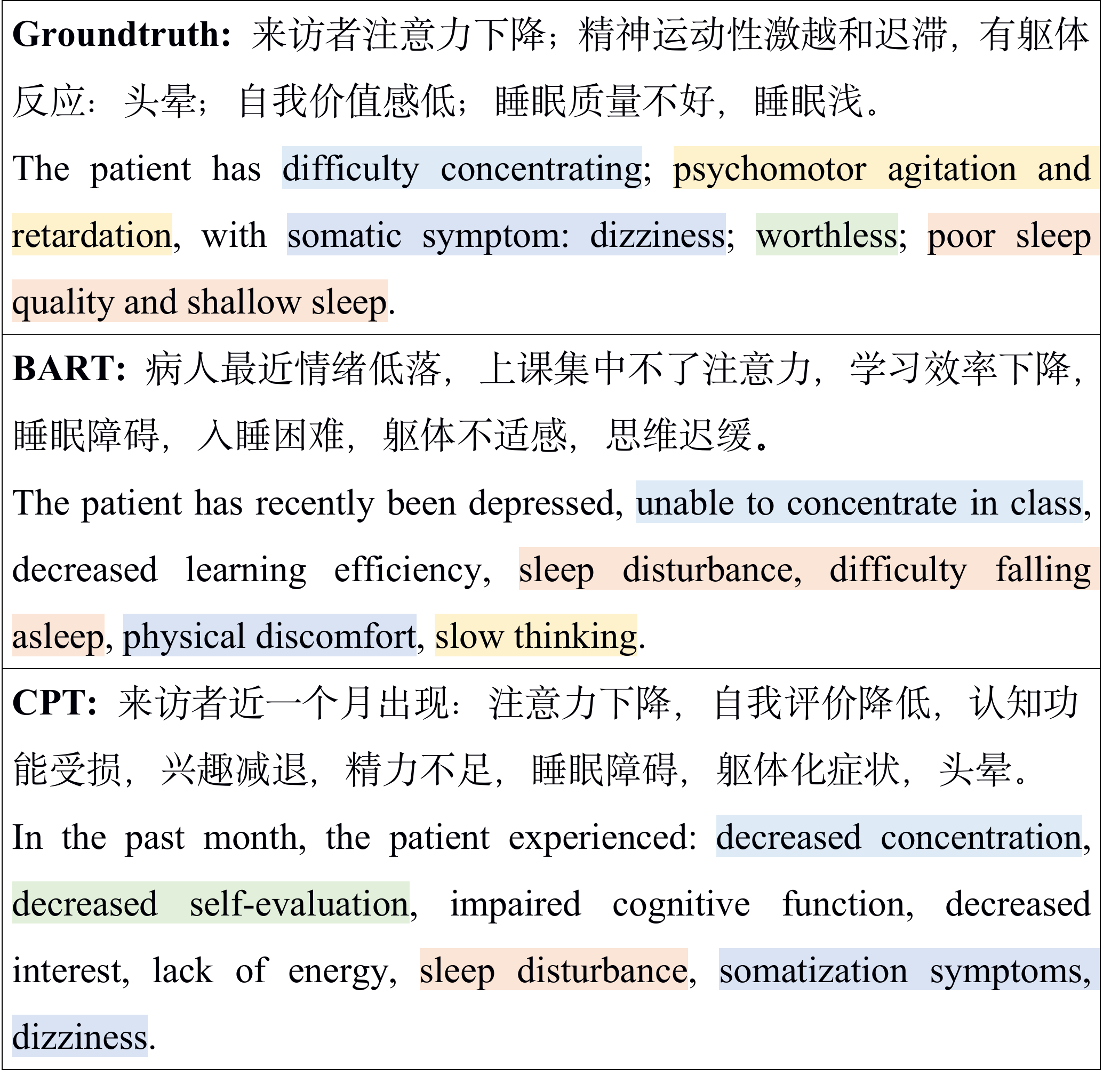}
    \caption{Examples of Dialog Summary Generation}
    \label{fig: Examples of generated summary}
\end{figure}

\begin{figure}[hbtp]
    \centering
    \includegraphics[width=0.48\textwidth,height=525pt]{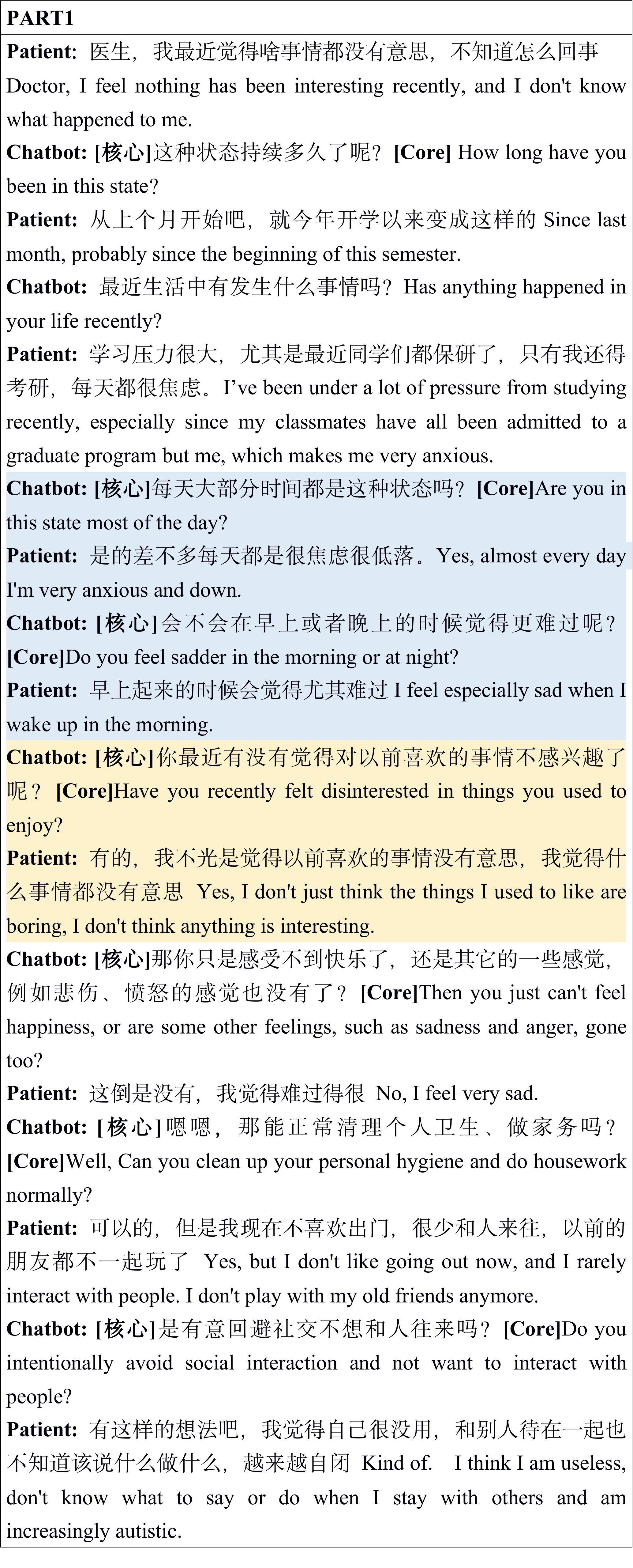}
    \caption{Human Interactive Example-part1}
    \label{fig: Human Interactive Evaluation Dialogue Example-PART1}
\end{figure}

\begin{figure}[t]
    \centering
    \includegraphics[width=0.48\textwidth,height=550pt]{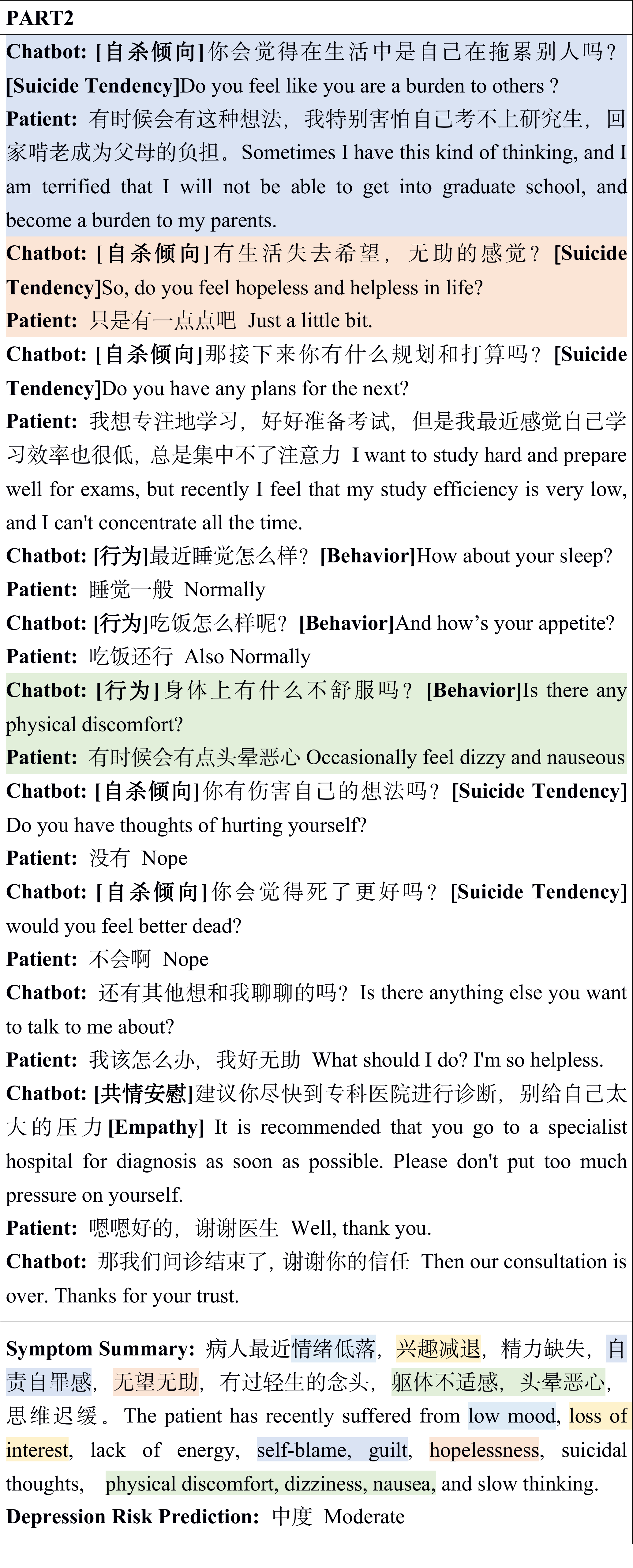}
    \caption{Human Interactive Example-part2}
    \label{fig: Human Interactive Evaluation Dialogue Example-PART2}
\end{figure}

\paragraph{Response Generation}
As shown in Figure \ref{fig: Examples of generated response}, we selected one representative example of the generated responses by different models. The examples in the figure show us that the correct topic helps the model generate more reliable and secure replies.
\paragraph{Dialog Summary Generation}In Figure \ref{fig: Examples of generated summary}, we present an example of the generated summary by different models. The models list most symptoms of the patient.
\paragraph{Human Interactive Example}
\label{ssec:Human Interactive Example}
We give a dialogue example with dialog summary and depressive severity generated by CPT during human evaluation in Figure \ref{fig: Human Interactive Evaluation Dialogue Example-PART1} and human evaluation in Figure \ref{fig: Human Interactive Evaluation Dialogue Example-PART2}.
In parentheses before the chatbot's sentence, we marked the topic predicted by the model. To clarify the correspondence between dialogue and summary, We have identified the correct symptom in the symptom summary with the same color as its location in the conversation. It can be seen that the model completed the entire consultation dialogue task and gave a dialogue summary covering almost all symptoms accurately.
\section{Worker Training Method}
\begin{figure*}
    \centering
    \includegraphics[width=\textwidth]{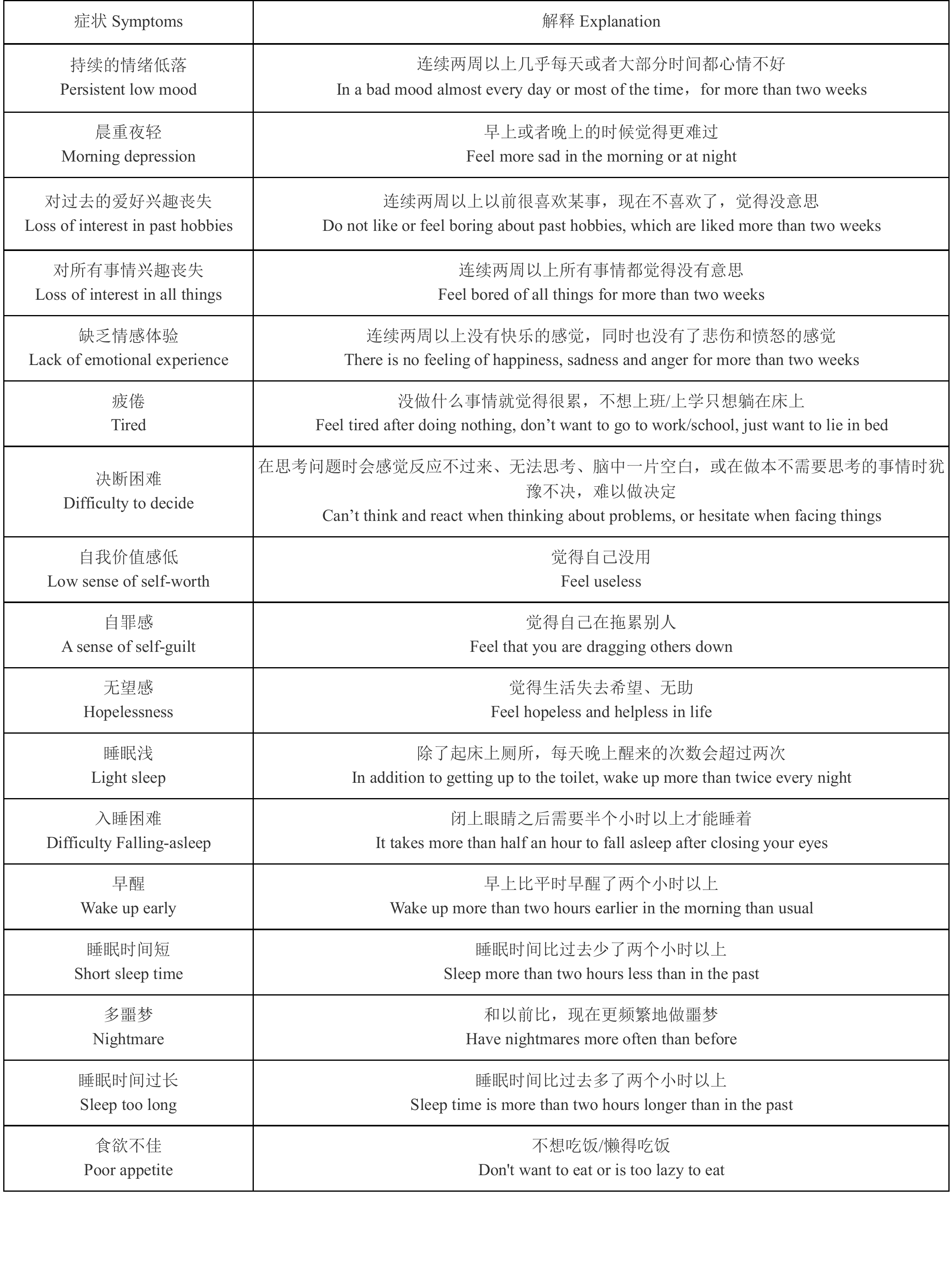}
    \caption{Explanation of Symptoms - 1}
    \label{fig: Explanation of Symptoms 1}
\end{figure*}

\begin{figure*}
    \centering
    \includegraphics[width=\textwidth]{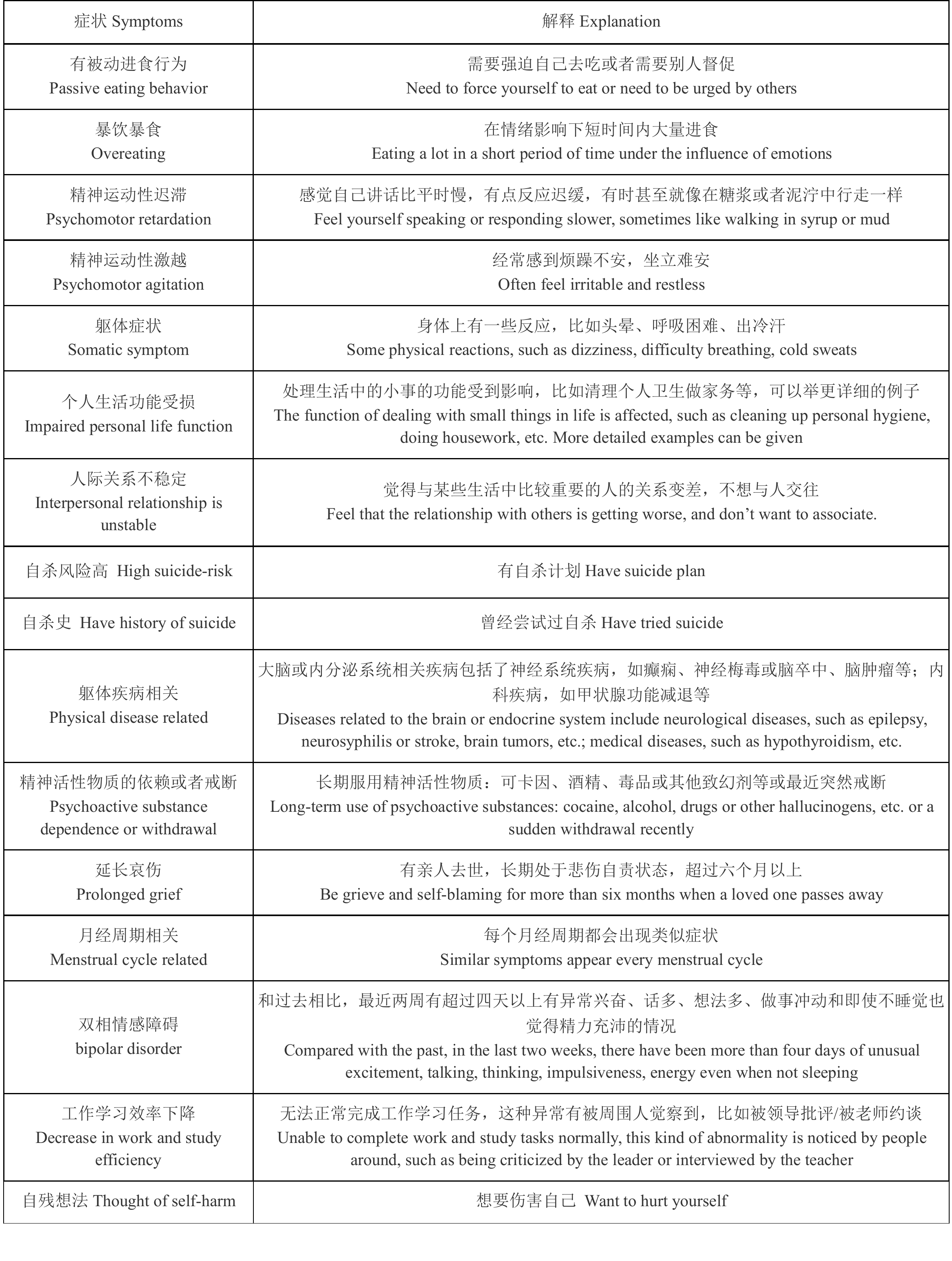}
    \caption{Explanation of Symptoms - 2}
    \label{fig: Explanation of Symptoms 2}
\end{figure*}

\begin{figure*}
    \centering
    \includegraphics[width=\textwidth]{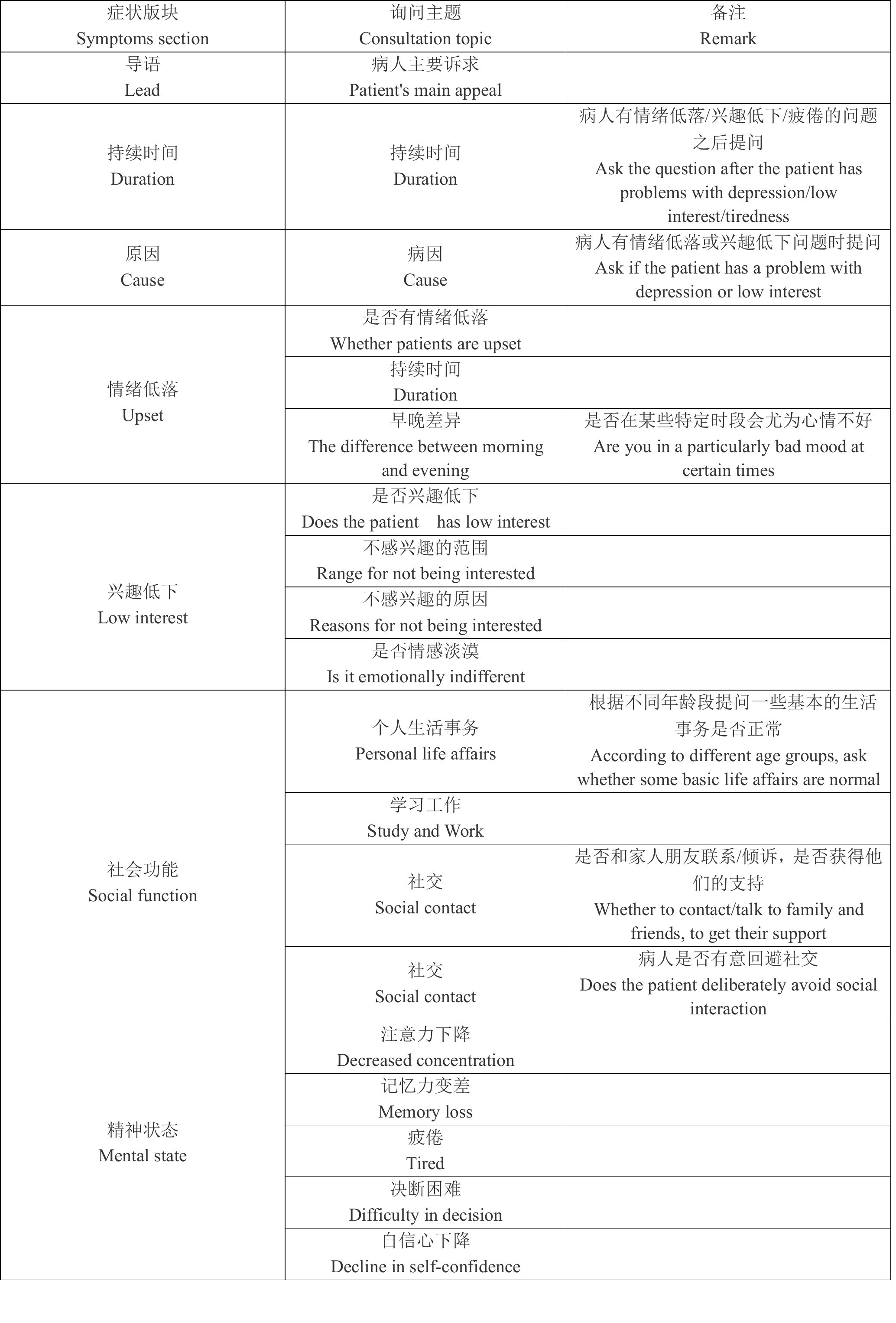}
    \caption{Doctors' questions - 1}
    \label{fig: Doctors' questions 1}
\end{figure*}

\begin{figure*}
    \centering
    \includegraphics[width=\textwidth]{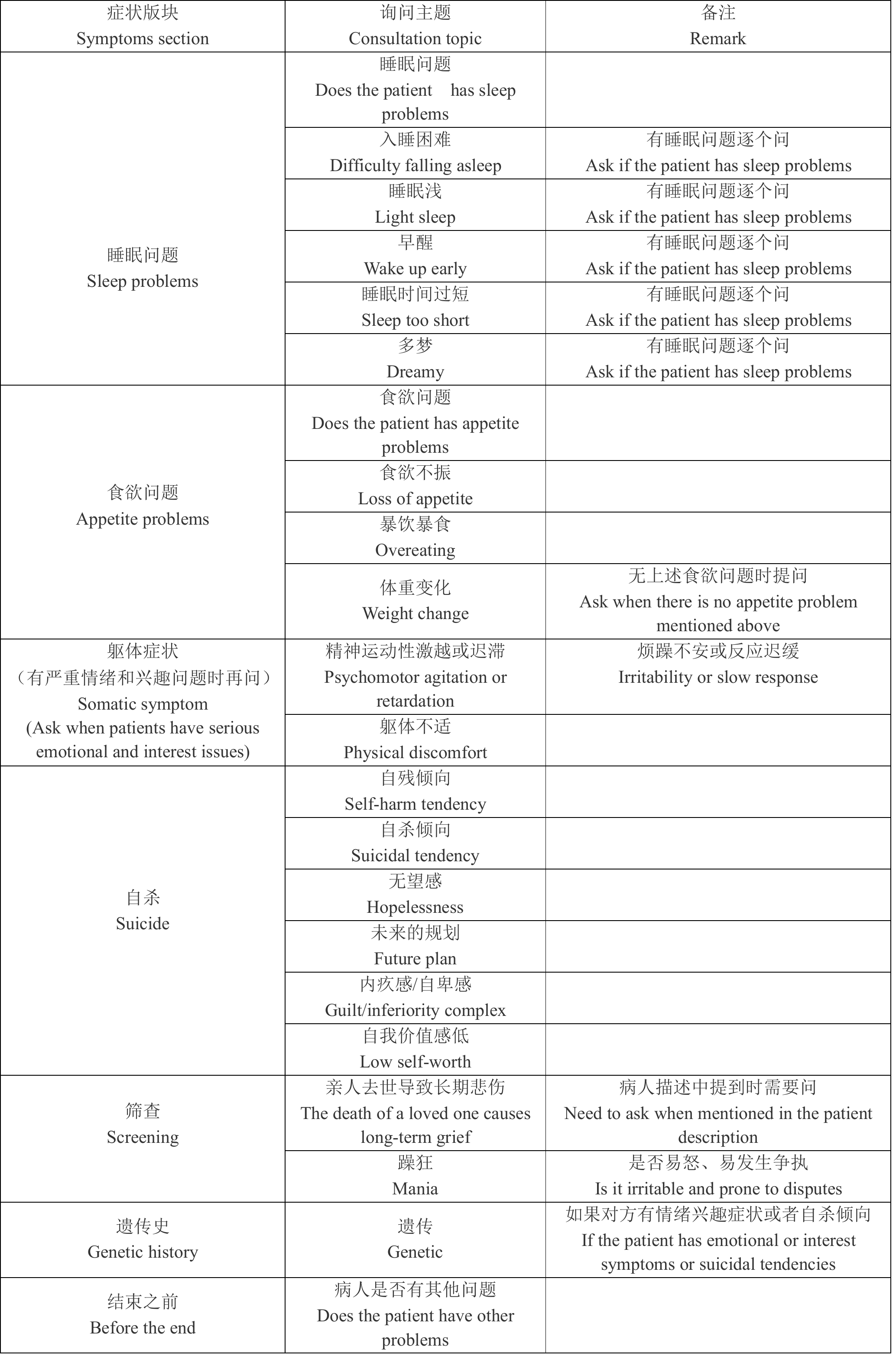}
    \caption{Doctors' questions - 2}
    \label{fig: Doctors' questions 2}
\end{figure*}

\paragraph{Acting Patients} To help acting patients better interpret the symptoms in the patient portraits, we provide detailed explanations of the symptoms in Figure \ref{fig: Explanation of Symptoms 1} and Figure \ref{fig: Explanation of Symptoms 2}, including the severity and duration. Besides expressing symptom accurately, they are required to imagine possible life events of the portrait's provider and talk with a doctor about it to express the patient's inner feelings in the process of telling the events.
\paragraph{Acting Doctors} We compile the 41 symptom items in Figure \ref{fig: Doctors' questions 1} and Figure \ref{fig: Doctors' questions 2} that doctors need to know when diagnosing depression, and design the questioning logic between questions of asking symptoms from mild to severe. The basic requirement is to obtain enough information from the patient during the conversation. At the same time, in order to further improve the dialogue experience, we require the acting doctors to: 1) Conduct the dialogue centered on the patient's complaint, that is, give priority to asking the patient's initiative symptom-related questions; 2) Ask further questions based on the patient's experience to elicit additional disclosure; 3) Give the patient certain feedback, e.g., empathy or comfort words when the patient talks about what they are going through.
\section{Quality Control}
To create transparency about quality control, the statistics of dialogues removed is provided in Table \ref{tab: Statistics of Removed Dialogues}. We have collected 4,457 dialogues, and 961 dialogues are removed because they haven't completed the entire diagnosis dialogues. 1,814 dialogues are automatically dropped by the stringent quality control criteria in Table \ref{tab:Quality Control Criteria}. Professional psychiatrists and clinical
psychotherapists screening the dialogues dropped 342 dialogues which unsuccessfully meet clinical standards. Eventually, we selected 1,339 dialogues into D$^4$.
\begin{table}[hbtp]\small
    \centering
    \setlength\tabcolsep{5pt} 
    \resizebox{1\linewidth}{!}{
    \begin{tabular}{cc}
    \hline
    \textbf{Reason for Removing}&\textbf{Sum} \\
    \hline
    Total& 4,457 \\
    \hline
    Unfinished& 961 \\
    \hline
    Dropped by Quality Control Criteria& 1,814 \\
    \hline
    Dropped by Doctor& 342 \\
    \hline
    Our Dataset& 1,339 \\
    \hline
    \end{tabular}
    }
    \caption{Statistics of Removed Dialogues}
    \label{tab: Statistics of Removed Dialogues}
\end{table}
\section{The Data Collection Platform}
\begin{figure*}
    \centering
    \includegraphics[width=\textwidth]{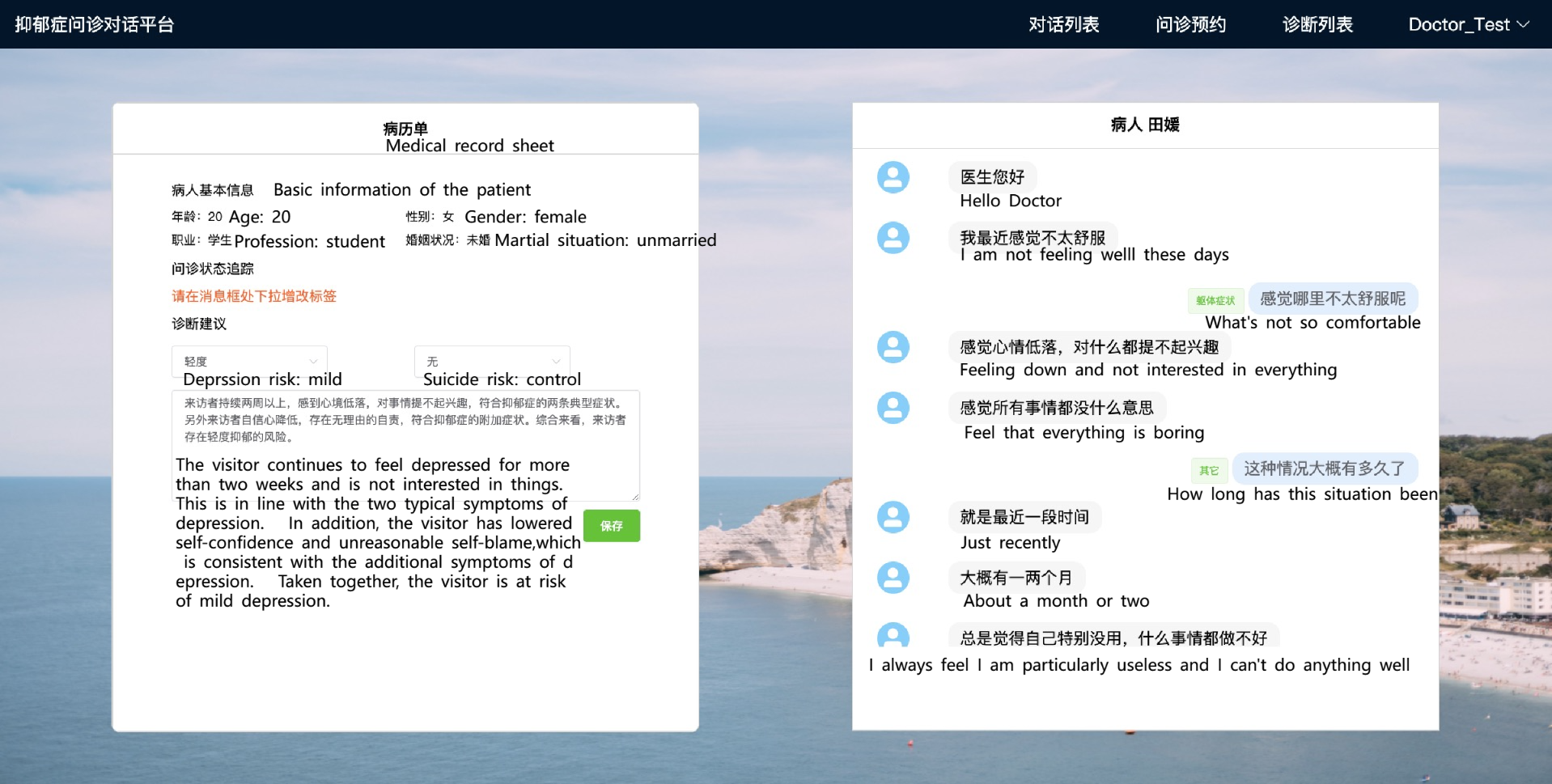}
    \caption{Page of doctor}
    \label{fig: Page of doctor}
\end{figure*}

\begin{figure*}
    \centering
    \includegraphics[width=\textwidth]{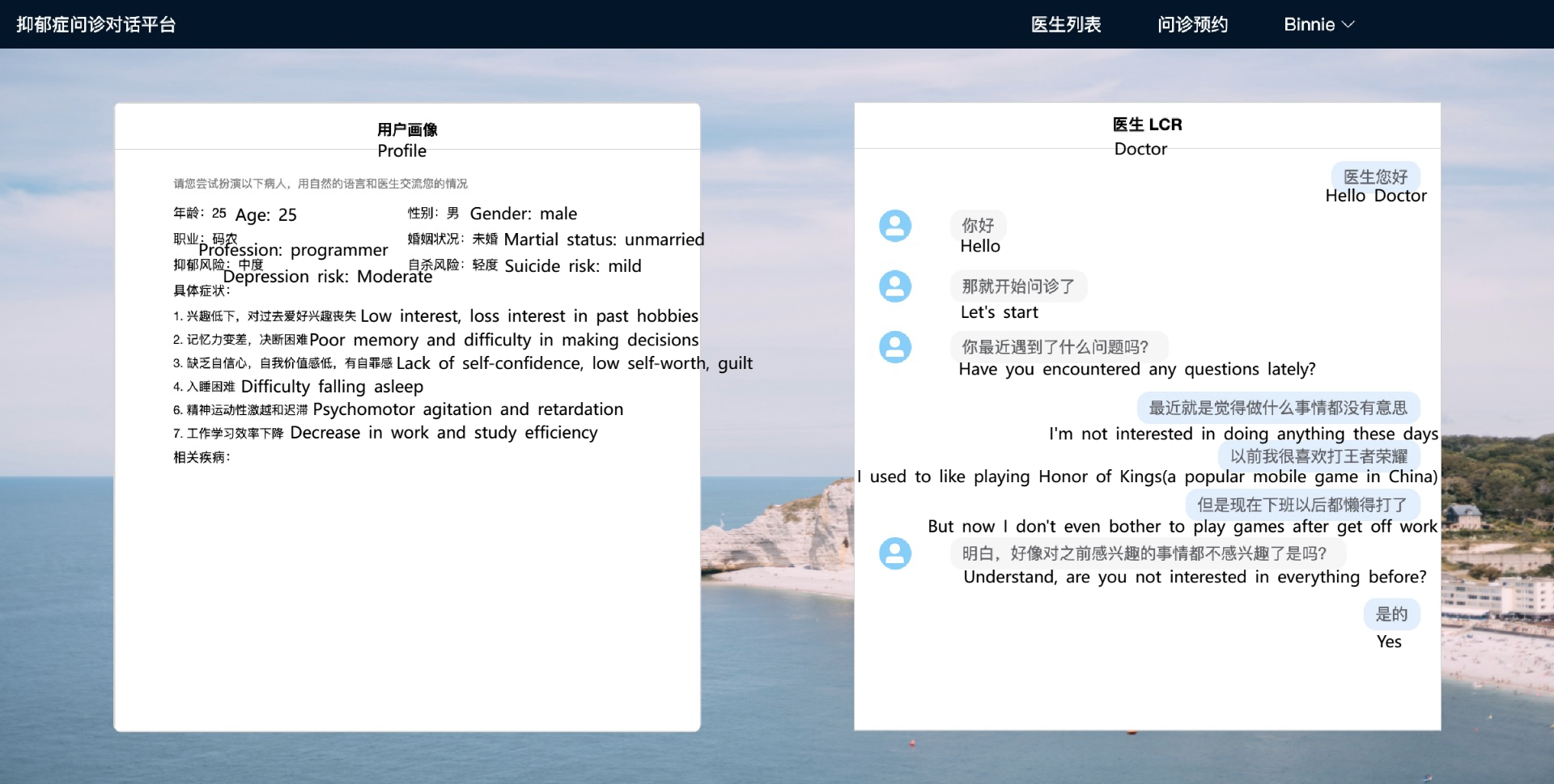}
    \caption{Page of patient}
    \label{fig: Page of patient}
\end{figure*}

Figure \ref{fig: Page of doctor} is screenshot of doctors' user interface, and Figure \ref{fig: Page of patient} is screenshot of the patients'.


\label{sec:appendix}
\end{document}